\documentclass[runningheads]{llncs}

\usepackage{eccv}

\usepackage{eccvabbrv}

\usepackage{graphicx}
\usepackage{booktabs}

\usepackage[accsupp]{axessibility}  % Improves PDF readability for those with disabilities.

\usepackage{mathrsfs}

\usepackage{hyperref}

\usepackage{orcidlink}

\usepackage{multirow}
\usepackage{colortbl}
\definecolor{title_gray}{gray}{.9}

\definecolor{wzt}{RGB}{255,0,0}
\definecolor{red}{RGB}{255,0,0}
\definecolor{green}{RGB}{0,255,0}
\definecolor{blue}{RGB}{0,0,255}
\definecolor{orange}{RGB}{255,165,0}

\begin{document}

\title{Enhancing 3D Lane Detection and Topology Reasoning with 2D Lane Priors} 

\titlerunning{Topo2D}

% Second Author\inst{2,3}\orcidlink{1111-2222-3333-4444} \and
% Third Author\inst{3}\orcidlink{2222--3333-4444-5555}}
\author{Han Li\inst{1} \and
Zehao Huang\inst{2} \and
Zitian Wang\inst{1} \and
Wenge Rong\inst{1} \and
Naiyan Wang\inst{2} \and
\\ Si Liu\inst{1}
}

\authorrunning{H.~Li et al.}

\institute{Beihang University \and TuSimple \\
\email{\{bryce18373631, zehaohuang18, a2394797795, winsty\}@gmail.com \\
\{w.rong, liusi\}@buaa.edu.cn}
}

\maketitle

\begin{abstract}
  3D lane detection and topology reasoning are essential tasks in autonomous driving scenarios, requiring not only detecting the accurate 3D coordinates on lane lines, but also reasoning the relationship between lanes and traffic elements. 
  Current vision-based methods, whether explicitly constructing BEV features or not, all establish the lane anchors/queries in 3D space while ignoring the 2D lane priors.
  In this study, we propose Topo2D, a novel framework based on Transformer, leveraging 2D lane instances to initialize 3D queries and 3D positional embeddings.
  Furthermore, we explicitly incorporate 2D lane features into the recognition of topology relationships among lane centerlines and between lane centerlines and traffic elements.
  Topo2D achieves 44.5\% OLS on multi-view topology reasoning benchmark OpenLane-V2 and 62.6\% F-Socre on single-view 3D lane detection benchmark OpenLane, exceeding the performance of existing state-of-the-art methods.
  The codes are released at \href{https://github.com/homothetic/Topo2D}{https://github.com/homothetic/Topo2D}.
\end{abstract}

% 各个任务的准确描述
% 3D Lane Detection
% Furthermore, we define the subtask of 3D lane detection as detecting directed 3D lane centerlines from the given multi-view images covering the whole horizontal FOV.

% Traffic Element Recognition
% on the given image in the front view, the location of traffic elements (traffic lights and road signs) and their attributes are demanded to be perceived simultaneously. Compared to typical 2D detection datasets, the challenge is that the size of traffic elements is tiny due to the large scale of outdoor environments.

% Topology Recognition
% Given multi-view images, the model learns to recognize the topology relationships among lane centerlines and between lane centerlines and traffic elements. 

\section{Introduction}

% \begin{figure}[t]
%   \centering
%   \begin{subfigure}{0.54\linewidth}
%     \centering
%     \includegraphics[width=\linewidth]{figs/figure1_a_final.pdf}
%     \caption{}
%     \label{fig:figonea}
%   \end{subfigure}
%   \hfill
%   \begin{subfigure}{0.44\linewidth}
%     \centering
%     \includegraphics[width=\linewidth]{figs/figure_1b.pdf}
%     \caption{}
%     \label{fig:figoneb}
%   \end{subfigure}
%   \caption{(a) 
%   % Comparison between different methods. 
%   Previous methods randomly initialize 3D lane queries in 3D space, while our method initializes 3D lane queries given 2D lane priors. (b) Comparison of lane detection recall under different thresholds. In both 2D recall and 3D recall, our model shows marked advancement relative to baseline MapTR\cite{Maptr} across various thresholds.}
%   \label{fig:figone}
% \end{figure}

\begin{figure}[t]
    \centering
    \begin{minipage}[b]{0.5\textwidth}
        \centering
        \subcaptionbox{\label{fig:figonea}}{
            \includegraphics[width=0.95\linewidth]{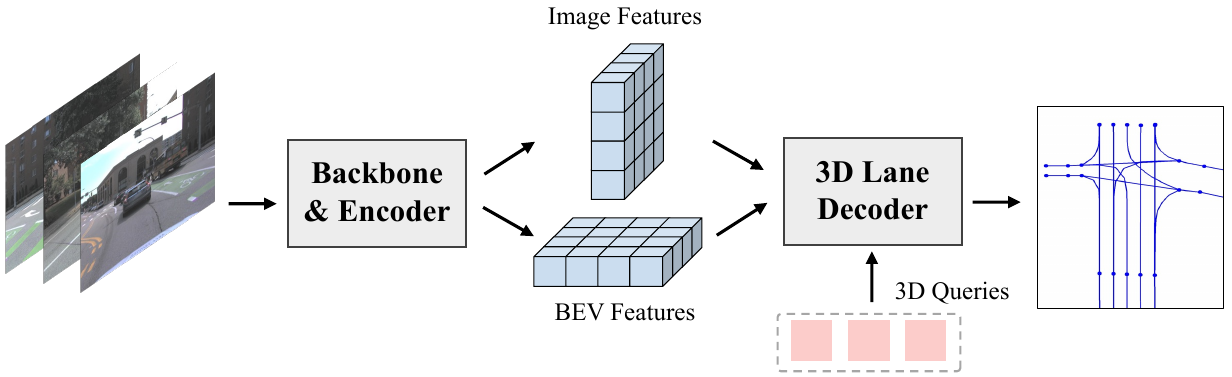}
        }
        \vfill
        \subcaptionbox{\label{fig:figoneb}}{
            \includegraphics[width=0.95\linewidth]{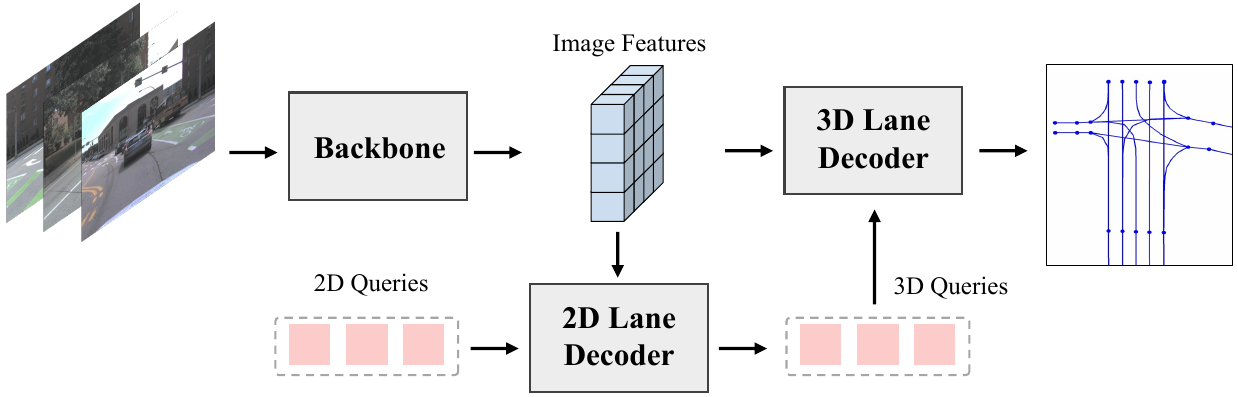}
        }        
    \end{minipage}
    \hfill
    \begin{minipage}[b]{0.45\textwidth}
        \centering
        \subcaptionbox{\label{fig:figonec}}{
            \includegraphics[width=0.95\linewidth]{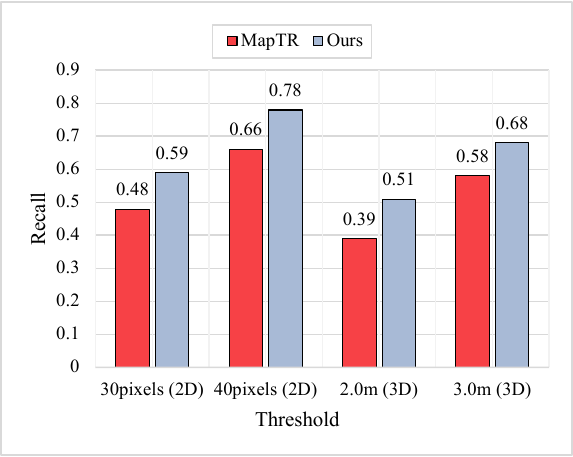}
        }
    \end{minipage}
    \caption{(a) Previous methods randomly initialize 3D lane queries in 3D space. (b) Our method initializes 3D lane queries given 2D lane priors. (c) Comparison of lane detection recall under different thresholds. In both 2D recall and 3D recall, our model shows marked advancement relative to baseline MapTR\cite{Maptr} across various thresholds.}
    \label{fig:figone}
\end{figure}

% 3d lane detection一直是自动驾驶领域的核心任务之一，对三维场景车道的检测，不仅可以指示可行驶区域，还可以支持运动预测，控制等下游任务。
% \textcolor{wzt}{first introduce lane detection, then state the advantage of 3D lane over 2D lane.}
3D lane detection, which focuses on detecting the accurate location of lanes, is one of the key components in the applications of the Advanced Driver Assistance System (ADAS). With the development and advancement of ADAS technology, the emergence of higher-level assisted driving functions, such as Navigate on Autopilot, has gradually shifted the demand for lane perception from 3D lane detection to the online High-Definition (HD) map construction\cite{pivotnet, Maptr, Hdmapnet, Vectormapnet, Maptrv2}. In addition to detecting lane segments, online HD map construction also requires the identification of other static traffic elements, such as traffic lights and road signs, and requires the topology between lane segments themselves as well as between lanes and traffic elements\cite{Openlane-v2, toponet}.

No matter for 3D lane detection task or online HD map construction, detecting 3D lanes is a critical component. Recent vision-based methods for lane detection can be mainly grouped into two categories. The first stream focuses on preserving 2D image features, leveraging interactions between predefined 3D lane anchors (or queries) and 2D features for the final predictions \cite{Anchor3dlane, TopoMLP, Petrv2}. For example, Anchor3DLane\cite{Anchor3dlane} initializes various 3D lane anchors and projects them onto 2D features using camera parameters, subsequently extracting features through bilinear interpolation. Other methods, such as MapTR \cite{Maptr}, explicitly construct bird's-eye-view (BEV) features \cite{toponet, Persformer, 3d-lanenet, Gen-lanenet, Maptr, bevlanedet}. They define 3D lane queries in the same BEV space and employ attention mechanisms \cite{deformdetr, Bevformer} for feature aggregation.

Though these methods have achieved remarkable performances, they all establish the lane anchors in 3D space while ignoring the 2D lane priors. In this paper, we propose a new framework that leverages 2D lane priors to enhance the performance of lane detection. As shown in \cref{fig:figonea} and \cref{fig:figoneb}, instead of adopting 3D queries initialized in 3D space, we utilize the 2D lane instances obtained by a 2D lane decoder as 3D queries. 
% \footnote{use subfigure and footnote. Don't write the subcaption in the figure directly. The fonts should also be consistent with other figures in the paper.}
This strategy is motivated by an observation: lane detection within the 2D image space generally results in a higher recall than detection in 3D. We conduct experiments on the OpenLane-V2\cite{Openlane-v2} dataset, comparing the recall of MapTR and our method in both 2D and 3D spaces. Our findings, illustrated in \cref{fig:figonec}, demonstrate that detecting 2D lanes in image space directly achieves 12\% (78\% vs. 66\%) higher recall than MapTR\footnote{For 2D recall, we project 3D Ground Truth and the prediction of MapTR into the image space for evaluation.}. By integrating 2D priors, our method achieves a superior 3D recall compared to MapTR.

The benefit of involving 2D lane priors is also proven in the task of topology reasoning. Existing methods\cite{toponet,TopoMLP} usually take two 3D queries and infer the topology structures through the concatenated features of them. 
% However, the topology between lanes and traffic elements are established in the front view, and the positional relationships of lanes in images can also provide additional information for establishing topology relationships between lanes.
% \footnote{I don't see the connection and logic of this sentence.} 
However, on the one hand, the positional relationships of 2D lanes provide additional information for establishing topological relationships between 3D lanes. On the other hand, since traffic elements are detected in images, the topology prediction of lanes and traffic elements requires the incorporation of 2D lane positions.
In our framework, we explicitly incorporate 2D lane features into the prediction of topology relationships.

Our framework, named Topo2D, enhances both lane detection and topology reasoning capabilities by integrating 2D lane priors. The 2D and 3D lane detectors are with similar transformer-based\cite{transformer} architectures. For 3D lane detection, we use the 2D lane query features and 2D coordinates obtained by the 2D lane decoder to initialize the 3D lane queries and positional embeddings. For topology prediction, we utilize a comprehensive approach that not only involves the features from 3D lanes and traffic elements, but also integrates corresponding 2D lane features, thereby enhancing overall performance.
In summary, our contributions can be outlined as follows:
\begin{itemize}
\item We propose to initialize 3D lane queries and positional embeddings using 2D lane priors, thereby enhancing 3D lane perception performance.
\item We explicitly utilize 2D lane information to assist the model in better recognizing the topology relationships among lane centerlines and between lane centerlines and traffic elements.
\item We validate our Topo2D on the multi-view topology reasoning benchmark OpenLane-V2\cite{Openlane-v2} and the single-view 3D lane detection benchmark OpenLane\cite{Persformer}. Topo2D achieves state-of-the-art performance on both benchmarks.
\end{itemize}

% We will release our code publicly upon acceptance.

\section{Related Work}

\subsection{Lane Detection}
% 车道线检测是自动驾驶场景理解的重要任务，基于锚点的方法因为简洁和高效，是二维车道检测任务中较为流行的方法。LineCNN首先定义了从图像边界发射的直线射线，以拟合2D车道线的形状，并应用非极大值抑制（NMS）来仅保留置信度较高的车道。LaneATT开发了基于锚点的特征汇集方法，用于提取2D锚点的特征。CLRNet通过特征金字塔学习迭代地对初始锚点进行细化。
The objective of 2D lane detection is to identify the precise location of lanes in 2D images. Among various methods \cite{qin2020ultra, linecnn, laneatt} developed for 2D lane detection, anchor-based approaches have risen to prominence for their simplicity and efficiency. 
% have risen popularity owing to their simplicity and efficiency. 
LineCNN\cite{linecnn} first proposes a novel representation method for lane anchors and achieving end-to-end lane detection.
Then, LaneATT\cite{laneatt} develops anchor-based attention mechanism to gather global information.
CLRNet\cite{clrnet} iteratively refines initial anchors through feature pyramids.
% 三维车道线检测需要根据二维图像输入，去预测车道准确的三维坐标。早期的三维车道线检测方法将图像的特征转换到bev视角，例如3DLaneNet，Gen-LaneNet，3D-LaneNet++，基于flat ground假设，使用IPM进行view transform，PersFormer使用cross attn建立图像特征和bev特征的关系，这些方法对于三维空间的垂直信息利用不足，在面临复杂的现实场景，比如上下坡的时候，检测精度会明显下降。

Building upon 2D lane detection, 3D lane detection predicts the 3D spatial coordinates of lanes. Early methods attempt to convert 2D image features into BEV features based on the assumption of flat ground. 
3D-LaneNet\cite{3d-lanenet} utilizes Inverse Perspective Mapping (IPM) to transform image features and then regresses lane segments on the resulted BEV feature map. 
Gen-LaneNet\cite{Gen-lanenet} introduces virtual top view based on IPM to achieve better perspective transformation. 

However, flat ground assumption fails when encountering uphill/downhill scenarios, which are common in real-world driving scenes. To remedy this issue, PersFormer\cite{Persformer} employs Perspective Transformer to construct more robust BEV features. 
% Anchor3DLane在3d空间定义anchor，并基于和图像特征的交互，直接预测三维车道线。LATR通过构建lane-aware query，并给图像添加dynamic 3d ground pe，以更好的基于2d图像提取三维特征，取得当前的最好结果。
% Recently, some methods directly regress the position of lane lines based on image features. 
Anchor3DLane\cite{Anchor3dlane} defines anchor lines in 3D space and aggregates image features of lanes based on camera parameters, directly predicting the 3D lane lines. LATR\cite{Latr} adds 3D positional embeddings to the images, constructing image features that include 3D information, and designs lane-aware queries to extract 3D features from the images.

\subsection{High Definition (HD) Map Construction}
% 在线高精地图构建可以看做三维车道线检测的扩展任务，除了检测车道线外还需要检测道路边界和人行横道等道路元素。现有的高精地图构建方法可以分为rasterized map和vectorized map两类，rasterized map方法（VPN CVT GKT）将高精地图构建建模为bev视角下的语义分割问题，但矢量化地图可以提供结构化的实例信息。
Multi-view-based online HD map construction can be regarded as an extension of the single-view-based 3D lane detection. In addition to detecting lanes, it also requires other static semantic elements in surrounding environment, such as pedestrian crossings and road boundaries.
Conventionally HD map is constructed offline with SLAM-based methods\cite{shan2018lego, shan2020lio}.
With the advancement in BEV representation learning, recent studies\cite{vpn, cvt, gkt} focus on predicting rasterized maps through BEV semantic segmentation. However, compared to vectorized maps, rasterized maps lack crucial instance-level information, such as lane structure, which is essential for downstream tasks.

% HDMapNet延续基于语义分割的思路，首先生成鸟瞰图视角特征图，然后通过后处理区分不同的实例，VectorMapNet将不规则的地图元素建模成一系列关键点，并通过Polyline Generator得到有序点集，MapTR MapTRV2进一步改进建模方法，将每个地图元素建模为具有一组等价排列的点集，并提出了分层查询将在线矢量化高精地图构建简化为并行回归问题。
For constructing vectorized HD maps, HDMapNet\cite{Hdmapnet} follows a semantic segmentation-based approach. After generating BEV feature maps and obtaining the segmentation results of map elements, HDMapNet distinguishes different instances through post-processing. 
To pursue an end-to-end approach, VectorMapNet\cite{Vectormapnet} models irregular map elements as a series of keypoints and predicts ordered point sets through a polyline generator. 
MapTR\cite{Maptr} and MapTRv2\cite{Maptrv2} represent each map element as a set of equivalently arranged points. They propose hierarchical queries to simplify the task into a parallel regression problem.

% compare our method
% 与这些在三维空间定义anchors或者queries检测车道线的方法不同（包括基于单视角图像的三维车道线检测方法和基于多视角图像的在线高精地图构建方法），我们的topo2d在二维空间定义queries，利用二维检测器提供的车道特征和车道坐标等先验，帮助三维检测器学习更多样的车道形状分布，以获得更好的三维检测性能。
Unlike these methods that define anchors or queries in 3D space for 3D lane perception, our Topo2D defines queries in 2D space and utilizes 2D priors such as lane features and lane coordinates to assist the 3D lane detector. 
% Due to the existence of 2D priors, the 3D lane detector learns a more diverse distribution of lane shapes, resulting in enhanced performance.
Thanks to the 2D priors, the 3D lane detector learns more comprehensive image features, resulting in enhanced performance.

\subsection{Topology Reasoning}
% 随着自动驾驶技术的发展，感知模型不仅需要精准检测场景中元素的位置，也需要对场景中元素的关系进行理解。自动驾驶场景拓扑理解需要识别车道之间，以及车道和交通元素之间的拓扑关系。TopoNet在分别检测centerline和traffic element之后，使用graph module对关系建模，利用相邻车道线和相关的交通元素，对车道的形状和位置进行refine。topomlp丢弃了复杂的图神经网络，而是使用简单的mlp进行topo关系理解，也取得了很好的效果。
With the introduction of the OpenLane-V2\cite{Openlane-v2} dataset, topology reasoning in driving scenes has attracted increasing attention.
This task involves recognizing the topologies among lanes and lanes with traffic elements (e.g., traffic lights and road signs).
TopoNet\cite{toponet} employs a graph module to model the relationships after separately detecting centerlines and traffic elements, refining the shape and position of lanes using adjacent lanes and relevant traffic elements. 
TopoMLP\cite{TopoMLP} abandons complex graph neural networks and instead employs simple Multilayer Perceptrons (MLPs) to predict the relationships.
Different from previous work, in order to achieve better consistency with more reliable 2D detection results, our Topo2D not only utilizes 3D lane features but also explicitly incorporates 2D lane features for topology relationship prediction.

% \subsection{Temporal Modeling}
% % 历史帧信息对于三维感知任务十分关键，显示构建bev特征的方法一般对bev特征进行时序的融合（BEVFormer BEVFusion），streampetr则是将上一帧和当前帧的sparse query一起送进decoder layer进行空间和时间的融合。
% Historical frame information plays a crucial role in the 3D perception tasks of autonomous driving scenes. Methods such as BEVFormer\cite{Bevformer} and BEVFusion\cite{Bevfusion} explicitly construct BEV features and directly fuse BEV features over time. StreamPETR\cite{streampetr} combines sparse queries from both the previous frame and the current frame into the decoder layer for spatial and temporal fusion.
% % 在车道线感知领域，Anchor3dlane将当前帧的anchor和历史帧的图像进行特征的交互，并对当前帧的预测进行多次refine，PETRv2将历史帧的feat和pos embed基于相机pose变换到当前帧，streammapnet使用MLP在latent space对query进行motion建模，并将历史帧的gt投影到当前帧对motion建模进行监督。
% In the field of lane detection, Anchor3DLane\cite{Anchor3dlane} interacts the anchors from the current frame with images from historical frames to refine the predictions of lanes in the current frame. PETRv2\cite{Petrv2} transforms features and positional embeddings from historical frames into the current frame based on camera pose. StreamMapNet\cite{Streammapnet} employs MLPs to model motion in the latent space and supervises motion modeling by projecting ground truth from historical frames onto the current frame.

\section{Method}

\begin{figure}[tb]
  \centering
  \includegraphics[width=\linewidth]{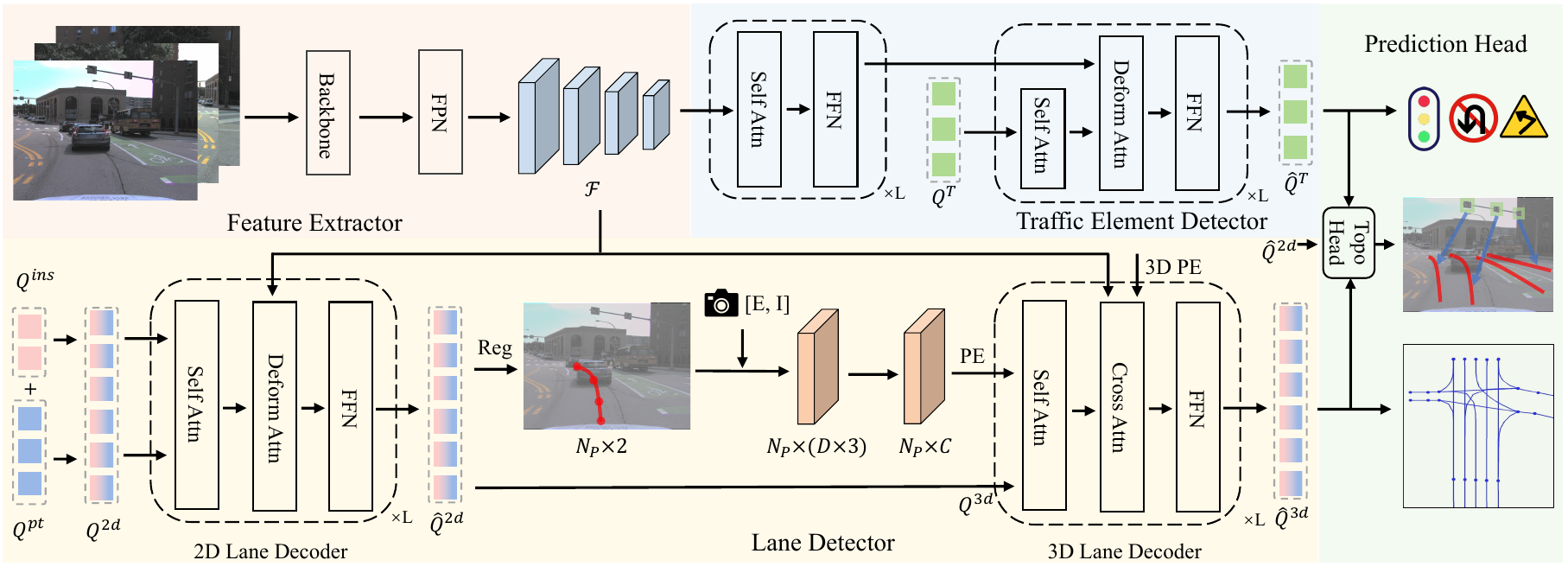}
  \caption{\textbf{The overall architecture of Topo2D.} Given multi-view images, 
  the images are first input to backbone network and FPN to extract image features. The image features are then fed into the subsequent 2D lane detector and 2D traffic element detector. The 3D lane detector initializes 3D lane queries and 3D position embeddings based on 2D lane priors and outputs 3D lane detection results. Finally, the 2D lane features are fused with the 3D lane features, and their relationships are estimated based on fused lane features and traffic element features.
  }
  \label{fig:pipeline}
\end{figure}

\subsection{Overview}
\cref{fig:pipeline} shows the overall architecture of the proposed Topo2D. Topo2D can solve multiple tasks within a single pipeline, including lane detection, traffic element detection, lane-to-lane topology prediction, and lane-to-traffic element topology prediction.

Given multi-view images $\mathcal{I}=\{\mathbf{I}_i\in \mathbb{R}^{3\times H_{I}\times W_{I}}|i=1,2,...,N_{I}\}$ from $N_{I}$ camera views, the images are input to the backbone network (e.g., ResNet-50\cite{resnet}) and FPN\cite{fpn} to extract multi-view multi-level 2D features $\mathcal{F}=\{\mathbf{F}_{i}^v\in \mathbb{R}^{C\times H_{F}^v\times W_{F}^v} | i=1,2,...,N_{I}; v=1,2,...,V\}$, where $V$ is the number of FPN levels.
% Topo2D accomplishes lane detection, traffic element detection and topology reasoning progressively. 
The lane detector consists of two stages, 2D lane detection and 3D lane detection, and finally outputs a fixed number of 3D lane instances $\{\mathbf{L}_{i}\in \mathbb{R}^{N_{P}\times 3}| i=1,2,...,l\}$, where $N_{P}$ is the point number of each instance. The traffic element detector is based on Deformable DETR\cite{deformdetr}, which generates 2D bounding boxes of traffic elements $\{\mathbf{T}_{i}\in \mathbb{R}^{4}|i=1,2,...,t\}$ from the front camera view. MLP-based topology prediction head receives features extracted by the detectors to predict the connectivity $\mathbf{G}^{ll}\in \mathbb{R}^{l\times l}$ among lanes and indication relationships $\mathbf{G}^{lt}\in \mathbb{R}^{l\times t}$ between traffic elements and lanes. The details will be described in the following sections.

\subsection{Lane Detector}
% 现有的基于Transformer结构的三维车道线检测器，例如maptr，一般直接在三维空间中随机初始化query，为了获得更高的模型召回率，我们设计使用基于二维车道特征的三维查询初始化方法，先在二维图像上对二维车道线进行检测，去学习车道更为多样和具有泛化性的形状分布。
% Existing transformer-based 3D lane detectors\cite{Persformer, Latr, Maptr, Maptrv2} typically initialize queries randomly in 3D space.
% \textcolor{wzt}{check the motivation}
% In Topo2D, we design a method for initializing 3D lane queries based on 2D lane priors.
% In order to learn a more diverse distribution of lane shapes and improve the detection recall rate in complex driving scenes, we design a method for initializing 3D queries based on 2D lane features.
In Topo2D,  the 2D lane detector 
% based on the Deformable DETR\cite{deformdetr} decoder 
is first applied to each image to localize lane instances in the perspective views.
% We use a 2D\_lane\_detection\_method ... .
Based on the detection results, we initialize 3D lane queries which comprise appearance features and positional embeddings derived from 2D lanes. 
The appearance features are generated from 2D lane features and the positional embeddings are initialized from 2D lane coordinates.
These lane-related geometric and semantic information help better associate relevant features in the subsequent self-attention and cross-attention modules.
% The queries and position embeddings from each view are combined and fed into the 3D lane decoder. 
The 3D lane detection part is built with a DETR-style\cite{detr} decoder, where the 3D lane queries interact with image features and output 3D lane coordinates after query updates. 
% Firstly, 2D lane detection is performed on the two-dimensional image to learn a more diverse and generalizable distribution of lane shapes.
% Then, we use the 2D detector to extract lane features to initialize queries for the 3D detector. Additionally, we employ 2D lane coordinates to initialize positional embeddings. 
% In the 3D lane detector, we apply a global cross-attention mechanism between the 3D queries initialized based on 2D features and the image features augmented with 3D coordinate information, to obtain more accurate 3D lane coordinates.

\subsubsection{2D Lane Detection.} 
We denote a 2D lane $\mathbf{L}_{i}^{2d}\in \mathbb{R}^{N_P\times2}$ as $N_P$ points along side the 2D lane. 
In each camera view, we define a set of instance-level queries $\{\mathbf{Q}^{ins}_i\}^{N_L}_{i=1}$ 
% where $l^{'}=\frac{l}{N_{I}}$ is the number of instance 
and a set of point-level queries $\{\mathbf{Q}^{pt}_j\}^{N_P}_{j=1}$, following the practice of MapTR\cite{Maptr}. So all lanes can be represented as a set of hierarchical queries $\mathbf{Q}^{2d}=\{\mathbf{Q}^{2d}_{ij}|i=1,...,N_L; j=1,...,N_P\}$, where:
\begin{equation}
  \mathbf{Q}^{2d}_{ij}= \mathbf{Q}^{ins}_i + \mathbf{Q}^{pt}_j.
  \label{eq:qij}
\end{equation}

% The decoder contains multiple layers for updating hierarchical queries iteratively. 
Within each 2D lane decoder layer, the hierarchical queries go through a self-attention module, a cross-attention module, and a feed-forward network sequentially to update features. 
In the self-attention module, we adopt vanilla multi-head attention\cite{transformer} to exchange information among hierarchical queries. In the cross-attention module, we adopt multi-scale deformable attention\cite{deformdetr} to aggregate information from extracted image features. This process can be formulated as:
\begin{gather}
\mathbf{X} = \text{LN}(\text{MultiHeadAttn}(\mathbf{Q}^{2d}, \mathbf{P}^{2d})) + \mathbf{Q}^{2d}, \\
\hat{\mathbf{X}} = \text{LN}(\text{MSDeformAttn}(\mathbf{X}, \mathbf{F})) + \mathbf{X}, \\
\hat{\mathbf{Q}}^{2d} = \text{LN}(\text{FFN}(\hat{\mathbf{X}})) + \hat{\mathbf{X}},
\end{gather}
where $\mathbf{P}^{2d}$ is 2D positional embedding as used in DETR\cite{detr}.
% and $\text{MultiHeadAttn}(\cdot)$, $\text{MSDeformAttn}(\cdot)$, $\text{LN}(\cdot)$, and $\text{FFN}(\cdot)$ denote multi-head attention, multi-scale deformable attention, layer normalization, and feed-forward network respectively.
% \textcolor{wzt}{add equations for self-attn and FFN}
% \begin{equation}
%   \hat{\mathbf{Q}}^{2d}_{ij} = \sum\limits_{m=1}^{M} \mathbf{W}_m(\sum\limits_{s=1}^S \sum\limits_{k=1}^K \mathbf{A}_{msk} \mathbf{W}_m^{'} \mathbf{F}^s[\phi_s(\text{MLP}(\mathbf{Q}^{2d}_{ij})) + \delta_{msk}]),
%   \label{eq:deformattn}
% \end{equation}
% where $m$ indexes the attention head, $s$ indexes the input feature level, and $k$ indexes the sampling point. $\delta_{msk}$ and $\mathbf{A}_{msk}$ denote the sampling offset and attention weight of the $k^{th}$ sampling point in the $s^{th}$ feature level and the $m^{th}$ attention head, respectively. Function $\phi_s$ in \cref{eq:deformattn} normalizes and re-scales the coordinates generated by MLPs to the input feature map of the $s^{th}$ level.

% The hierarchical query predicts the 2D coordinates $P_{ij} = (x_{ij}, y_{ij})$ of the lane point in the image space. 
% Each lane $L_{i}$ corresponds to a set of points $\{P_{ij}\}_{j=0}^{N_{L}-1}$ with flexible and dynamic distribution. 
The prediction heads consist of a classification branch and a point regression branch. The classification branch predicts the probability of the lane instance belonging to a specific class:
\begin{equation}
\label{eq:clshead}
\mathbf{S}_{i}^{2d} = \text{Linear}(\text{AvgPooling}(\{\hat{\mathbf{Q}}^{2d}_{ij}\}_{j=1}^{N_P})),
\end{equation}
% The point regression branch predicts the positions of the point. 
and the point regression branch predicts the 2D coordinates of lane points  $\mathbf{L}_{i}^{2d}=\{\mathbf{L}^{2d}_{ij}\}_{j=1}^{N_P}\in \mathbb{R}^{N_P\times2}$ in the image space: 
% \textcolor{wzt}{add equations}
% For each lane in, it outputs an $N_{L}\times C$ dimensional feature vector representing aggregated image features, as well as an $N_{L}\times 2$ dimensional vector representing normalized 2D coordinates.
% \textcolor{wzt}{add equations}
\begin{equation}
\label{eq:reghead}
\mathbf{L}_{ij}^{2d} = \text{Linear}(\hat{\mathbf{Q}}^{2d}_{ij}).
\end{equation}

\subsubsection{3D Lane Query Initialization.}
\label{sec:queryinitial}
% The 3D lane detector is inspired by the 3D multi-view object detector PETR\cite{petr}. PETR\cite{petr} retains image features and encodes the position information of 3D coordinates into image features, enabling the object queries to interact with 3D position-aware image features.
We directly use the updated 2D lane features $\hat{\mathbf{Q}}^{2d}$ after the 2D lane decoder as the appearance features $\mathbf{Q}^{3d}$ of the 3D lane queries. Additionally, we generate 3D positional embeddings based on the predicted 2D lane coordinates.
Specifically, 
% to elucidate the connection between the 2D lane and the 3D spatial domain, 
we use a camera ray to represent each 2D lane point in the 3D coordinate system. 
Following PETR\cite{petr}, the camera rays are discretized to generate a set of points, where each point can be represented as  $\mathbf{p}_{ijk}^{cam} = (u_{ij}\times d_k, v_{ij}\times d_k, d_k)$. 
$(u_{ij},v_{ij})$ is the 2D lane point coordinate in the image space and $\{d_k\}_{k=1}^D$ are the pre-defined depth values along the axis orthogonal to the image plane. 
% We further calculate the corresponding 3D coordinates in 3D world coordinate system and normalize the 3D points based on the perception ranges. 
$\mathbf{p}_{ijk}^{cam}$ in the camera coordinate system is then transformed to  $\mathbf{p}_{ijk}$ in the world coordinate system using camera parameters.
Finally, we concatenate $\mathbf{p}_{ijk}$ and feed them into an MLP to produce 3D positional embeddings:
\begin{equation}
  \mathbf{P}_{ij}^{3d}=\text{MLP}(\text{Concat}(\{\mathbf{p}_{ijk}\}_{k=1}^D)).
  \label{eq:3dpe}
\end{equation}
% where $D$ is the point number in each ray.

\subsubsection{3D Lane Detection.} 
% In the 3D lane detection section, to enable better localization of queries in the image, we add positional embeddings to each pixel on the image using a similar approach to the one used for 2D lane points \textcolor{wzt}{(what does it mean?)}. 
% Combining the 2D lane features and 2D lane coordinates from multiple views, we obtain a series of hierarchical 3D lane queries $\{\mathbf{Q}^{3d}_{ij}|i=1,2,...,N_L\times N_I;j=1,2,...,N_P\}$ and their corresponding 3D positional embeddings $\{\mathbf{P}_{ij}^{3d}|i=1,2,...,N_L\times N_I;j=1,2,...,N_P\}$.
Given the appearance features $\mathbf{Q}^{3d}$ and positional embeddings $\mathbf{P}^{3d}$ of 3D lane queries, the 3D lane detector iteratively refines the appearance features.  
The structure of 3D lane detector is similar to the 2D counter part, but the cross-attention module is implemented with a global multi-head attention to aggregate 3D lane features:
% from 3D position-aware image features \textcolor{wzt}{(not defined)}
\begin{gather}
\mathbf{Y} = \text{LN}(\text{MultiHeadAttn}(\mathbf{Q}^{3d}, \mathbf{P}^{3d})) + \mathbf{Q}^{3d}, \\
\hat{\mathbf{Y}} = \text{LN}(\text{MultiHeadAttn}(\mathbf{Y}, \mathbf{F})) + \mathbf{Y}, \\
\hat{\mathbf{Q}}^{3d} = \text{LN}(\text{FFN}(\hat{\mathbf{Y}})) + \hat{\mathbf{Y}}.
\end{gather}
% where $\text{MultiHeadAttn}(\cdot)$, $\text{LN}(\cdot)$, $\text{FFN}(\cdot)$ denote multi-head attention, layer normalization, and feed-forward network respectively.
% \begin{equation}
%   \hat{\mathbf{Q}}^{3d}_{ij} = \sum\limits_{m=1}^{M} \mathbf{W}_m(\sum\limits_{n=1}^{N_{I}} \sum\limits_{s=1}^S \sum\limits_{k\in\Omega_{n}^s} \mathbf{A}_{mnsk} \mathbf{W}_m^{'} \mathbf{F}^s_n[k]),
%   \label{eq:globalattn}
% \end{equation}
% where $m$ indexes the attention head, $s$ indexes the input feature level, and $n$ indexes the camera view. $\Omega_{n}^s$ denotes all points on the feature map $\mathbf{F}^s_n$ and $\mathbf{A}_{mnsk}$ denotes the attention weight of the $k^{th}$ point in the $s^{th}$ feature level, the $n^{th}$ camera view and the $m^{th}$ attention head. 
Then, similar to \cref{eq:clshead} and \cref{eq:reghead}, the prediction heads predict instance class score $\mathbf{S}_{i}^{3d}$ and 3D coordinates $\mathbf{L}_{i}^{3d}=\{\mathbf{L}_{ij}^{3d}\}_{j=1}^{N_P}\in \mathbb{R}^{N_P\times3}$ based on $\hat{\mathbf{Q}}^{3d}$.
% \textcolor{wzt}{add equations}

\subsection{Topology Reasoning}

\begin{figure}[tb]
  \centering
  \includegraphics[width=0.8\linewidth]{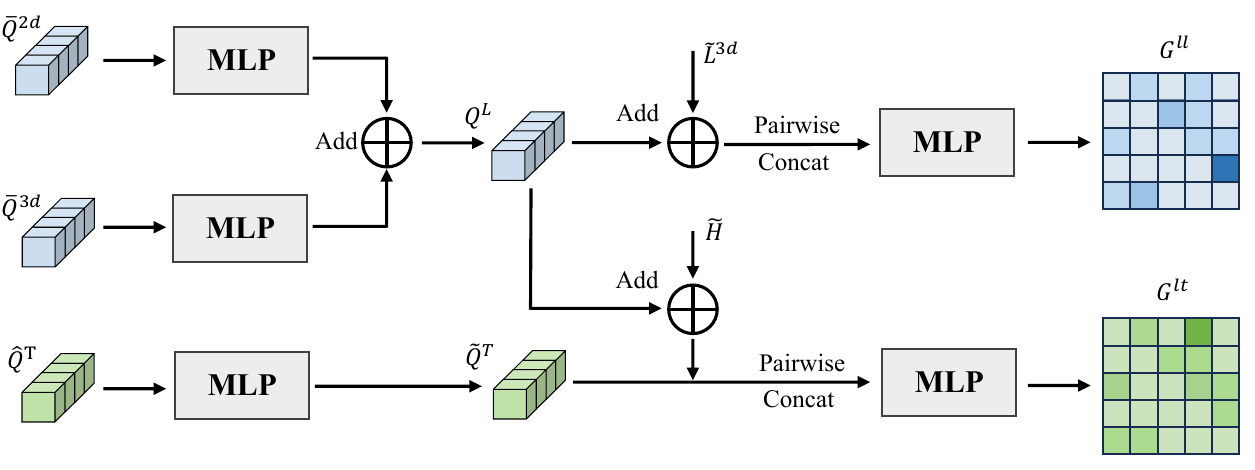}
  \caption{\textbf{Illustration of topology prediction heads.} 
  % The 2D lane query features and 3D lane query features are first fused by an addition operation, then
  First, we embed the 2D lane instance queries using MLPs, and add them with the embedded 3D lane instance queries. 
  % Then concatenate the lane queries with lane queries, and the lane queries with traffic element queries in pairs to predict the topology relationships. 
  Then we concatenate the lane queries with each other in pairs, as well as the lane queries with the traffic element queries in pairs, to predict the topology relationships.
  % Additionally, for the prediction of lane-traffic element relationships, we incorporate embeddings of camera parameters.
  Additionally, for lane-lane topology, we incorporate embeddings of the 3D coordinates of lane points, while for lane-traffic element topology, we incorporate embeddings of camera parameters.
  }
  \label{fig:topohead}
\end{figure}

We use deformable DETR\cite{deformdetr} as the traffic element detector to output bounding boxes and corresponding features $\{\hat{\mathbf{Q}}^{T}_i\}_{i=1}^{t}$ of the traffic elements in the front view. 
After obtaining lane and traffic element detection results,  2D lane information is explicitly incorporated in this process to boost the performance of topology inference.

On one hand, 2D lane segments significantly contribute to the inference of the connectivity among lane instances. On the other hand, since traffic elements are inherently captured in the image space, 2D lane information facilitates more accurate alignments with traffic element information. 

Specifically, we first utilize average pooling operation to convert hierarchical queries into instance queries $\{\Bar{\mathbf{Q}}_{i}^{3d}\}_{i=1}^l$ and $\{\Bar{\mathbf{Q}}_{i}^{2d}\}_{i=1}^l$, where $l=N_I\times N_L$. After that, we embed the 2D lane queries using MLPs and add them with embed 3D lane queries:
% \textcolor{wzt}{where is 2D position used?}
\begin{equation}
  \mathbf{Q}_i^L= \text{MLP}(\Bar{\mathbf{Q}}_{i}^{3d}) + \text{MLP}(\Bar{\mathbf{Q}}_{i}^{2d}).
  \label{eq:add2dquery}
\end{equation}
Then, we also embed the 3d lane coordinates $\mathbf{L}^{3d}_{i}$ using MLPs:
\begin{equation}
    \tilde{\mathbf{L}}^{3d}_{i} = \text{MLP}(\text{Flatten}(\mathbf{L}^{3d}_{i})).
\end{equation}

As illustrated in \cref{fig:topohead}, the lane-lane topology reasoning branch predicts the connection among 3D lanes: 
\begin{equation}
  \mathbf{G}^{ll}_{mn}= \text{MLP}(\text{Concat}(\mathbf{Q}^L_m + \tilde{\mathbf{L}}^{3d}_m; \mathbf{Q}^L_n + \tilde{\mathbf{L}}^{3d}_n)),
  \label{eq:lltopo}
\end{equation}
where $\mathbf{Q}^L_m$, $\tilde{\mathbf{L}}^{3d}_m$ and $\mathbf{Q}^L_n$, $\tilde{\mathbf{L}}^{3d}_n$ denote the features and coordinate embeddings of the $m^{th}$ and $n^{th}$ lane. The lane-traffic element topology reasoning branch predicts the topology relationship between 3D lanes and traffic elements: 
\begin{gather}
  % \tilde{\mathbf{Q}}^{T}_n = \text{MLP}(\hat{\mathbf{Q}}^{T}_n),\\
  \tilde{\mathbf{H}}=\text{MLP}(\text{Flatten}(\mathbf{H}[:3, :])), \\
  \mathbf{G}^{lt}_{mn}= \text{MLP}(\text{Concat}(\mathbf{Q}^L_m + \tilde{\mathbf{H}}; \text{MLP}(\hat{\mathbf{Q}}^{T}_n))),
  \label{eq:lttopo}
\end{gather}
% \textcolor{wzt}{explain the meaning of symbols}
where $\mathbf{H}$ denotes a standard $4\times 4$ transformation matrix between 3D world coordinate system and 2D image coordinate system. $\mathbf{Q}^L_m$ and $\hat{\mathbf{Q}}^T_n$ denote the features of the $m^{th}$ lane and $n^{th}$ traffic element, respectively.
% \footnote{It is wired to transform the camera parameters using MLP then added to features. Why not predict them directly in 2D space?}

\subsection{Loss Function}
Our total loss function is defined as follows:
% \textcolor{wzt}{conflict usage of symbols. see Sec. 3.1.}
\begin{equation}
  \mathcal{L}_{\text{total}} = \mathcal{L}_{\text{2d-lane}} + \mathcal{L}_{\text{3d-lane}} + \mathcal{L}_{\text{t}} + \mathcal{L}_{\text{topo-ll}} + \mathcal{L}_{\text{topo-lt}},
  \label{eq:loss}
\end{equation}
where $\mathcal{L}_{\text{2d-lane}}$ and $\mathcal{L}_{\text{3d-lane}}$ are lane detection losses. Both of them adopt the focal loss\cite{focal} for classification and $\mathcal{L}_1$ loss for lane points regression. 
% In order to adapt to the arbitrary shape of 2D lane element, we further add an edge direction loss\cite{Maptr} to $L_{\text{2d}}$ to supervise the geometrical shape in the higher edge level.
$\mathcal{L}_{\text{t}}$ is traffic element detection loss, including focal loss, $\mathcal{L}_1$ loss and GIoU loss. The lane-lane topology loss $\mathcal{L}_{\text{topo-ll}}$ and lane-traffic topology loss $\mathcal{L}_{\text{topo-lt}}$ are focal losses for binary classification. We omit the weight for each loss for brevity.
% Since our method is a query-based method, it requires the matching between the predictions and ground-truth. In this work, we only use bipartite matching on the basic lane and traffic element detection. The matching is directly used in topology reasoning loss as well.
The details are presented in the supplementary materials.

\section{Experiments}

\subsection{Datasets and Metrics}
To evaluate the proposed method, we conduct experiments on two benchmarks: the multi-view topology reasoning benchmark OpenLane-V2\cite{Openlane-v2} and the single-view 3D lane detection benchmark OpenLane\cite{Persformer}.

% OpenLanev2是自动驾驶场景的大型perception and reasoning dataset，基于argo2和nuscene，每个subset包含2Hz的1k场景，eval指标包括centerline map，traffic element map，lane-lane topo map和lane-traffic map，以及这些指标的平均OpenLaneScore。
\subsubsection{OpenLane-V2}\cite{Openlane-v2} is a large-scale perception and reasoning dataset for driving scenes.
% \textcolor{wzt}{(topology reasoning?)}. 
It comprises two subsets, \textit{subset\_A} and \textit{subset\_B}, developed from Argoverse2\cite{argoverse} and nuScenes\cite{nuscenes} respectively. We utilize only \textit{subset\_A} due to the lack of height information in \textit{subset\_B}. OpenLane-V2 \textit{subset\_A} comprises 1000 scenes annotated at 2Hz, with each frame containing images from 7 views.

The evaluation metrics consist of four components: 
$\text{DET}_l$ and $\text{DET}_t$ measure the instance-level performance of centerline and traffic element using mean average precision (mAP). Specifically, $\text{DET}_l$ employs the Frech{\'e}t distance to quantify similarity, averaging over matching thresholds set at \{1.0m, 2.0m, 3.0m\}, while $\text{DET}_t$ employs Intersection over Union (IoU) and computes the average over different traffic categories. 
For topology reasoning metrics, $\text{TOP}_{ll}$ and $\text{TOP}_{lt}$ are mAP metrics adapted from the graph domain\footnote{The metrics for $\text{TOP}$ scores have been updated. We use the updated metrics in our  evaluation as suggested by \href{https://github.com/OpenDriveLab/OpenLane-V2/blob/master/docs/metrics.md}{OpenLane-V2}.}.
To evaluate the overall effect of detection and topology reasoning, the OpenLane-V2 Score (OLS) is computed as follows:
\begin{equation}
  \text{OLS} = \frac{1}{4}[\text{DET}_l+\text{DET}_t+f(\text{TOP}_{ll})+f(\text{TOP}_{lt})],
  \label{eq:OLS}
\end{equation}
where $f$ is a scale function to emphasize the task of topology reasoning. 

In addition, to better evaluate the lane detection results and compare them with previous lane detection methods, OpenLane-V2\cite{Openlane-v2} also provides a lane related metric $\text{DET}_{l,chamfer}$, which utilizes the Chamfer distance as the similarity measure and computes average over matching thresholds set at \{0.5m, 1.0m, 1.5m\}.

% OpenLane是基于waymo的大型三维车道线检测数据集，包含1k个场景和200k帧，车道线有14个类别，eval的指标包括f-score, acc, x error close/far, z error close/far。
\subsubsection{OpenLane}\cite{Persformer} is a 3D lane detection benchmark constructed upon the Waymo open dataset\cite{Waymo}. This dataset comprises 1000 segments, consisting of 200K frames. Each frame contains only front view images with a resolution of $1280\times 1920$. OpenLane includes 880K lane annotations distributed across 14 categories. 

During the evaluation process, predictions and ground-truths are matched via minimum cost flow, where pairwise costs are defined as the square root of the sum of squared point-to-point distances. A prediction is considered true positive (TP) if more than 75\% of the predicted points are within a distance threshold (i.e., 1.5m) from the ground-truth points. Based on this definition, the maximum F1 score is further computed, and x/z errors are calculated separately for near distance (0-40 meters) and far distance (40-100 meters). Additionally, category accuracy is reported, which computes the proportion of correctly predicted categories to all TP predictions.

\begin{table}[tb]
  \caption{\textbf{Comparison on topology reasoning task} on OpenLane-V2 \textit{subset\_A}. 
  The reported results of state-of-art methods are from TopoNet\cite{toponet}. 
  % The best is in \textbf{bold}.
  }
  \label{tab:openlanev2sota}
  \centering
  \begin{tabular}{@{}l|cc|c|cccc@{}}
    \toprule 
    Method & Backbone & Epoch & $\text{OLS}\uparrow$ & $\text{DET}_l\uparrow$ & $\text{DET}_t\uparrow$ & $\text{TOP}_{ll}\uparrow$ & $\text{TOP}_{lt}\uparrow$ \\
    \midrule
    % STSU\cite{stsu} & ResNet-50 & 24e & 25.4 & 12.7 & 43.0 & 0.5 & 15.1 \\
    % VectorMapNet\cite{Vectormapnet}$\;$ & ResNet-50 & 24e & 20.8 & 11.1 & 41.7 & 0.4 & 5.9 \\
    % MapTR\cite{Maptr} & ResNet-50 & 24e & 26.0 & 17.7 & 43.5 & 1.1 & 10.4 \\
    % % paper
    % TopoNet\cite{toponet} & ResNet-50 & 24e & 35.6 & 28.5 & 48.1 & 4.1 & 20.8 \\
    % % github
    % TopoNet\cite{toponet} & ResNet-50 & 24e & 35.6 & 28.6 & 48.6 & 4.1 & 20.3 \\
    % % paper
    % TopoMLP\cite{TopoMLP} & ResNet-50 & 24e & 38.2 & 28.3 & 50.0 & 7.2 & 22.8 \\
    % % github
    % TopoMLP\cite{TopoMLP} & ResNet-50 & 24e & 38.2 & 28.5 & 49.5 & 7.2 & 23.4 \\
    % \hline
    % % paper
    % \rowcolor{title_gray} TopoMLP\cite{TopoMLP} & Swin-B & 24e & 42.2 & 30.7 & 54.3 & 9.5 & 28.3 \\
    % \rowcolor{title_gray} TopoMLP\cite{TopoMLP} & Swin-B & 48e & 43.3 & 30.0 & 55.8 & 9.4 & 31.7 \\
    % \hline
    STSU\cite{stsu} & ResNet-50 & 24e & 29.3 & 12.7 & 43.0 & 2.9 & 19.8 \\
    VectorMapNet\cite{Vectormapnet}$\;$ & ResNet-50 & 24e & 24.9 & 11.1 & 41.7 & 2.7 & 9.2 \\
    MapTR\cite{Maptr} & ResNet-50 & 24e & 31.0 & 17.7 & 43.5 & 5.9 & 15.1 \\
    TopoNet\cite{toponet} & ResNet-50 & 24e & 39.8 & 28.6 & 48.6 & 10.9 & 23.8 \\
    Topo2D (Ours) & ResNet-50 & 24e & \textbf{44.5} & \textbf{29.1} & \textbf{50.6} & \textbf{22.3} & \textbf{26.2} \\
  \bottomrule
  \end{tabular}
\end{table}

\begin{table}[tb]
   \caption{\textbf{Comparison on centerline detection task} without incorporating traffic elements on OpenLane-V2 \textit{subset\_A}.
   % \emph{Topology} indicates whether the model is jointly trained with the lane-lane topology reasoning task. 
   }
  \label{tab:centerlinetopo}
  \centering
  \begin{tabular}{@{}l|c|cc@{}}
    \toprule 
    Method & Trained w/. Topo & $\text{DET}_l\uparrow$ & $\text{DET}_{l, chamfer}\uparrow$ \\
    \midrule
    VectorMapNet\cite{Vectormapnet}$\;$ & $\times$ & 12.7 & 10.3 \\
    MapTR\cite{Maptr} & $\times$ & 10.0 & 21.7 \\
    Topo2D (Ours) & $\times$ & \textbf{26.6} & \textbf{31.0} \\
    \hline
    STSU\cite{stsu} & $\checkmark$ & 14.2 & 13.8 \\
    TopoNet\cite{toponet} & $\checkmark$ & 27.7 & 27.4 \\
    Topo2D (Ours) & $\checkmark$ & \textbf{28.8} & \textbf{32.4} \\
  \bottomrule
  \end{tabular}
\end{table}

\subsection{Implementation Details}
We use ResNet-50\cite{resnet} as backbone networks, followed by FPN\cite{fpn} to generate multi-level features.
In the lane detector, we set 20 instance queries and 11 point queries in each camera view. Both the 2D and 3D decoders contain 6 decoder layers. The 2D lane ground-truths are obtained by projecting 3D lane ground-truths to each view and cropping the visible parts of each lane. The loss weights for classification and regression are set to 2.0 and 5.0, respectively.
% \textcolor{wzt}{use symbols defined above}
We use Deformable DETR\cite{deformdetr} as the traffic element detector. The loss weights for classification, regression, and GIoU are set to 2.0, 5.0 and 2.0. 
% \textcolor{wzt}{use symbols defined above}
Both the topology reasoning branches comprise a three-layer MLP, with a loss weight set to 5.0.
We use AdamW\cite{adamw} optimizer with a weight decay rate of 0.01, and use the cosine annealing policy\cite{cos} for learning rate adjustment.  
HSV augmentation and grid mask\cite{gridmask} are used during training. 
All the experiments are conducted on eight A100 GPUs. 
% For the OpenLane-V2 dataset\cite{Openlane-v2}, the front view images are first cropped and padded to match the size of other views. Then all images are resized to $775\times 1024$ with a scaling factor of 0.5. 
% % Then all images are resized with a scaling factor of 0.5 \textcolor{wzt}{resolution}. 
% The batch size is 8 with an initial learning rate of 3e-4.
% For the OpenLane\cite{Persformer} dataset, all input images are resized to $800\times 1024$. The structure of the lane detection model is similar to the centerline detection model used on OpenLane-V2\cite{Openlane-v2}. The batch size is 32 with an initial learning rate of 2e-4. 
More details are given in the supplementary materials.

\begin{table}[t]
  \caption{\textbf{Comparison with state-of-art methods} on OpenLane validation set. \emph{Cate-Acc} represents category accuracy. 
  % The best is in \textbf{bold}.
  }
  \label{tab:openlane}
  \centering
  \begin{tabular}{@{}l|cc|cccc@{}}
    \toprule 
    Method & F-Score$\uparrow$ & Cate-Acc$\uparrow$ & X-Near$\downarrow$ & X-Far$\downarrow$ & Z-Near$\downarrow$ & Z-Far$\downarrow$ \\
    \midrule
    3D-LaneNet\cite{3d-lanenet} & 44.1 & - & 0.479 & 0.572 & 0.367 & 0.443 \\
    Gen-LaneNet\cite{Gen-lanenet} & 32.3 & - & 0.593 & 0.494 & 0.140 & 0.195 \\
    PersFormer\cite{Persformer} & 50.5 & 89.5 & 0.319 & 0.325 & 0.112 & 0.141 \\
    CurveFormer\cite{Curveformer} & 50.5 & - & 0.340 & 0.772 & 0.207 & 0.651 \\
    Anchor3DLane\cite{Anchor3dlane}$\;$ & 53.7 & 90.9 & 0.276 & 0.311 & 0.107 & 0.138 \\
    BEVLaneDet\cite{bevlanedet} & 58.4 & - & 0.309 & 0.659 & 0.244 & 0.631 \\
    PETRv2\cite{Petrv2} & 61.2 & - & 0.400 & 0.573 & 0.265 & 0.413 \\
    LATR\cite{Latr} & 61.9 & 92.0 & \textbf{0.219} & 0.259 & \textbf{0.075} & \textbf{0.104} \\
    Topo2D (Ours) & \textbf{62.6} & \textbf{94.0} & 0.226 & \textbf{0.244} & 0.088 & 0.111 \\
  \bottomrule
  \end{tabular}
\end{table}

% \begin{table}[tb]
%   \caption{}
%   \label{tab:openlanef1}
%   \centering
%   \begin{tabular}{@{}l|>{\centering}p{1.0cm}|>{\centering}p{1.0cm}|>{\centering}p{1.0cm}|>{\centering}p{1.25cm}|>{\centering}p{1.0cm}|>{\centering}p{1.75cm}|c@{}}
%     \toprule 
%     & & Up \& & & Extreme & & & Merge \\
%     \multirow{-2}*{Method} & \multirow{-2}*{All} & Down & \multirow{-2}*{Curve} & Weather & \multirow{-2}*{Night} & \multirow{-2}*{Intersection} & \& Split \\
%     \midrule
%     3D-LaneNet\cite{3d-lanenet} & 44.1 & 40.8 & 46.5 & 47.5 & 41.5 & 32.1 & 41.7 \\
%     Gen-LaneNet\cite{Gen-lanenet} & 32.3 & 25.4 & 33.5 & 28.1 & 18.7 & 21.4 & 31.0 \\
%     PersFormer\cite{Persformer} & 50.5 & 42.4 & 55.6 & 48.6 & 46.6 & 40.0 & 50.7 \\
%     CurveFormer\cite{Curveformer} & 50.5 & 45.2 & 56.6 & 49.7 & 49.1 & 42.9 & 45.4 \\
%     Anchor3DLane\cite{Anchor3dlane}$\;$ & 53.7 & 46.7 & 57.2 & 52.5 & 47.8 & 45.4 & 51.2 \\
%     BEVLaneDet\cite{bevlanedet} & 58.4 & 48.7 & 63.1 & 53.4 & 53.4 & 50.3 & 53.7 \\
%     LATR\cite{Latr} & 61.9 & 55.2 & \textbf{68.2} & 57.1 & 55.4 & 52.3 & 61.5 \\
%     Ours & \textbf{62.6} & \textbf{55.5} & 67.7 & \textbf{59.1} & \textbf{57.4} & \textbf{52.4} & \textbf{62.5} \\
%   \bottomrule
%   \end{tabular}
% \end{table}

\subsection{Comparison on OpenLane-V2 Dataset}
We first compare the performance on the topology reasoning task. \cref{tab:openlanev2sota} shows the results on OpenLane-V2 \textit{subset\_A}. Our Topo2D achieves 44.5\% OLS when using ResNet-50 for training 24 epochs, surpassing other state-of-the-art methods. In terms of topology related metrics, compared to TopoNet, we achieve a 11.4\% improvement in $\text{TOP}_{ll}$ and a 2.4\% improvement in $\text{TOP}_{lt}$.
% It is worth noting that our 3D lane query initialization method not only enhances performance in lane detection (29.1\% vs. TopoNet 28.6\% on $\text{DET}_l$), but also significantly assists in predicting lane topology structures (22.3\% vs. TopoNet 10.9\% on $\text{TOP}_{ll}$)\footnote{we can not conclude the performances improvement comes from the 3D lane query initialization method}. \textcolor{wzt}{give an explanation}
% wzt: 再想想咋解释，这才是实验的重点。你统计过训练和测试时的预测数量没有
% With a more powerful backbone (i.e. Swin-B\cite{swin}) and more training time (i.e. 48 epochs), the OLS reaches 44.5.

It is noteworthy that some previous multi-view 3D lane detection methods choose to use Chamfer distance to evaluate the performance of unordered lane detection\cite{Vectormapnet, Maptr}.
To have a more comprehensive comparison, we also provide the comparison of centerline detection task under the same evaluation protocol, as marked by $\text{DET}_{l,chamfer}$ in \cref{tab:centerlinetopo}. 
% The results of previous methods are all from TopoNet\cite{toponet}. 
Topo2D achieves an improvement of 10.7\% on $\text{DET}_{l,chamfer}$ compared to MapTR when training without topology prediction and an improvement of 5.0\% compared to TopoNet with topology reasoning.

% 我们进一步研究了我们的方法为什么在det-lc上获得了比det-l上更高的提升。det-lc是基于点集的指标，偏向测算点到线段的距离，det-l是基于有向线段的指标，偏向计算点到点的距离，对于车道线的每个点尤其是起始点的位置非常敏感。对于较远处的车道线，由于车道线在2d上所占的像素点很少，我们的方法对于起点终点的预测相对不准确，导致我们的模型检测到了这些车道线，但无法和gt match，导致在det-l上的提升相对不明显。
We further investigate why our method achieves a higher improvement on $\text{DET}_{l,chamfer}$ compared to $\text{DET}_l$.
$\text{DET}_l$ is based on Fr{\'e}chet distance, treating the lane as a directed line, while $\text{DET}_{l,chamfer}$ uses the Chamfer distance, treating the lane as a set of points. $\text{DET}_l$ is more sensitive to the position of each point, especially the starting point and the ending point. 
For instance, distant lanes (\textcolor{orange}{orange} box in \cref{fig:centerlineb}) often occupy fewer pixels in the 2D image, which can cause predicted 3D lanes that align well with ground truths except for less precise starting/ending points. For those lanes, a Fr{\'e}chet distance-based metric is easier to treat them as false positives compared to a Chamfer distance-based metric.

% 另外，我们在200m的长距离检测上验证了我们模型的性能。相比于需要构建bev特征的baseline，我们的方法在检测的距离大幅度延长之后，性能相比于maptr有很大的提升。在添加了时序，使用历史帧的车道查询检测当前帧的车道之后，模型的性能进一步提升。
% Additionally, we validate the performance of Topo2D in long range detection. In \cref{tab:centerlinetopo}, compared to the baseline MapTR, our method exhibited significant performance improvements (19.9\% vs. 8.6\% on $\text{DET}_l$). With the utilization of lane queries from historical frames, the performance of Topo2D-T is further enhanced (+3.1\% on $\text{DET}_l$).

\subsection{Comparison on OpenLane Dataset}
% 我们在OpenLane的验证集上，将我们的方法和之前的方法进行比较，可以发现，在相似的车道检测error的情况下，我们的fscore和类别的预测精度分别超过latr 0.7%和2.0%。另外，我们在不同场景下比较了方法的fscore，我们的方法在极端环境和夜晚的检测效果提升较为明显。
We present the main results on OpenLane validation set in \cref{tab:openlane}. Compared to the state-of-the-art method LATR, Topo2D achieves an improvement of 0.7\% in F-Score and 2.0\% in category accuracy with similar x/z errors ($\pm$1.5cm), demonstrating its performance in accurately detecting 3D lanes.
We also provide the performance comparison under different scenarios in the supplementary materials.
% Topo2D outperforms LATR across most scenarios. 
% Specifically, we observe that our model performs more accurate lane detection under scenes where 2D lane features are difficult to identify, such as Extreme Weather and Night. This is due to our 2D lane queries extracting more comprehensive image information, allowing the 3D queries based on 2D lane prior initialization to better locate lane feature positions in the 3D lane detection decoder, achieving more accurate predictions of lane point positions.
% \textcolor{wzt}{why?}.

\subsection{Ablation Study}
In this section, we conduct ablation experiments on OpenLane-V2 \textit{subset\_A}. All the models are trained for 24 epochs with ResNet-50\cite{resnet} as backbone network. 
For modules specifically related to lane detection, we conduct experiments on centerline topology task without incorporating traffic elements, and compare the results on $\text{DET}_l$, $\text{DET}_{l,chamfer}$ and $\text{TOP}_{ll}$.
% \textcolor{wzt}{don not understand...}.

\begin{table}[tb]
    \caption{\textbf{Ablation studies} on OpenLane-V2 \textit{subset\_A}. All the experiments use ResNet-50 as backbone and train for 24 epochs.}
    \label{tab:ablation}
    \centering
    \begin{subtable}[tb]{0.50\linewidth}
        \centering
            \begin{tabular}{@{}cc|ccc@{}}
            \toprule 
            2D & Rand. & $\text{DET}_l\uparrow$ & $\text{DET}_{l, chamfer}\uparrow$ & $\text{TOP}_{ll}\uparrow$  \\
            \midrule
            & $\checkmark$ & 24.7 & 28.8 & 19.1 \\
            $\checkmark$ & & \textbf{28.8} & \textbf{32.4} & \textbf{22.4} \\
            $\checkmark$ & $\checkmark$ & 26.8 & 31.7 & 22.0 \\
            \bottomrule
            \end{tabular} 
        \caption{\textbf{Initialization method for 3D queries.} 
        % \emph{Rand.} denotes random initialization.
        }
        \label{tab:3dqueryinitial}
    \end{subtable}
    \begin{subtable}[tb]{0.49\linewidth}
        \centering
            \begin{tabular}{@{}c|ccc@{}}
            \toprule 
            Num. & $\text{DET}_l\uparrow$ & $\text{DET}_{l, chamfer}\uparrow$ & $\text{TOP}_{ll}\uparrow$  \\
            \midrule
            10 & 21.4 & 25.9 & 12.9 \\
            20 & \textbf{28.8} & \textbf{32.4} & \textbf{22.4} \\
            30 & 27.8 & 31.8 & 21.3 \\
            \bottomrule
            \end{tabular} 
        \caption{\textbf{Number of instance queries.} }
        \label{tab:querynum}
    \end{subtable}
        \begin{subtable}[tb]{0.50\linewidth}
        \centering
            \begin{tabular}{@{}c|ccc@{}}
            \toprule 
            Method & $\text{DET}_l\uparrow$ & $\text{DET}_{l, chamfer}\uparrow$ & $\text{TOP}_{ll}\uparrow$  \\
            \midrule
            2D & \textbf{28.8} & \textbf{32.4} & \textbf{22.4} \\
            3D & 23.8 & 27.9 & 17.6 \\
            \bottomrule
            \end{tabular} 
        \caption{\textbf{Design of 2D lane ground truths.}}
        \label{tab:2dsample}
    \end{subtable}
    \begin{subtable}[tb]{0.49\linewidth}
        \centering
            \begin{tabular}{@{}c|ccc@{}}
            \toprule 
            Query & $\text{DET}_l\uparrow$ & $\text{DET}_{l, chamfer}\uparrow$ & $\text{TOP}_{ll}\uparrow$  \\
            \midrule
            Ins. & 25.8 & 30.1 & 21.2 \\
            Hie. & \textbf{28.8} & \textbf{32.4} & \textbf{22.4}  \\
            \bottomrule
            \end{tabular} 
        \caption{\textbf{Design of 3D lane queries.}}
        \label{tab:ptsquery}
    \end{subtable}
    \begin{subtable}[tb]{1.0\linewidth}
        \centering
            \begin{tabular}{@{}ccc|c|cccc@{}}
            \toprule 
            3D & Proj. & 2D & $\text{OLS}\uparrow$ & $\text{DET}_l\uparrow$ & $\text{DET}_t\uparrow$ & $\text{TOP}_{ll}\uparrow$ & $\text{TOP}_{lt}\uparrow$ \\
            \midrule
            $\checkmark$ & & & 43.8 & 28.6 & \textbf{50.9} & 21.5 & 24.5 \\
            $\checkmark$ & $\checkmark$ & & 44.1 & 28.8 & 50.7 & 21.8 & 25.3 \\
            $\checkmark$ & $\checkmark$ & $\checkmark$ & \textbf{44.5} & \textbf{29.1} & 50.6 & \textbf{22.3} & \textbf{26.2} \\
            \bottomrule
            \end{tabular} 
        \caption{\textbf{Impact of adding 2D information.} 
        % \emph{Proj.} denotes projection matrix.
        }
        \label{tab:2dquerytopo}
    \end{subtable}
\end{table}

\begin{figure}[t]
  \centering
  \begin{subfigure}{0.45\linewidth}
    \centering
    % \flushright
    \includegraphics[width=\linewidth]{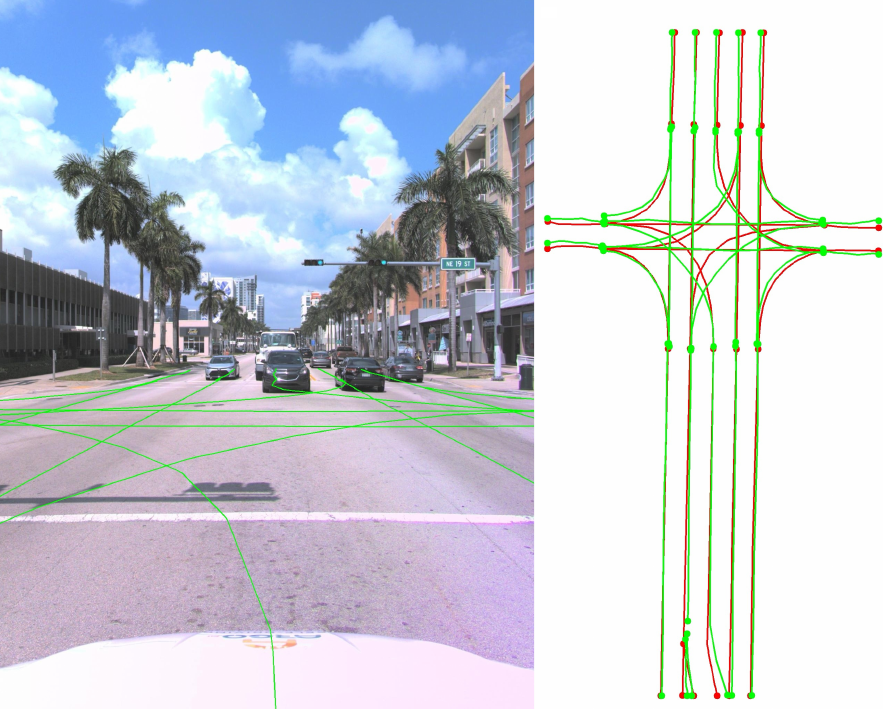}
    \caption{}
    \label{fig:centerlinea}
  \end{subfigure}
  % \hfill
  \begin{subfigure}{0.45\linewidth}
    \centering
    % \flushleft
    \includegraphics[width=\linewidth]{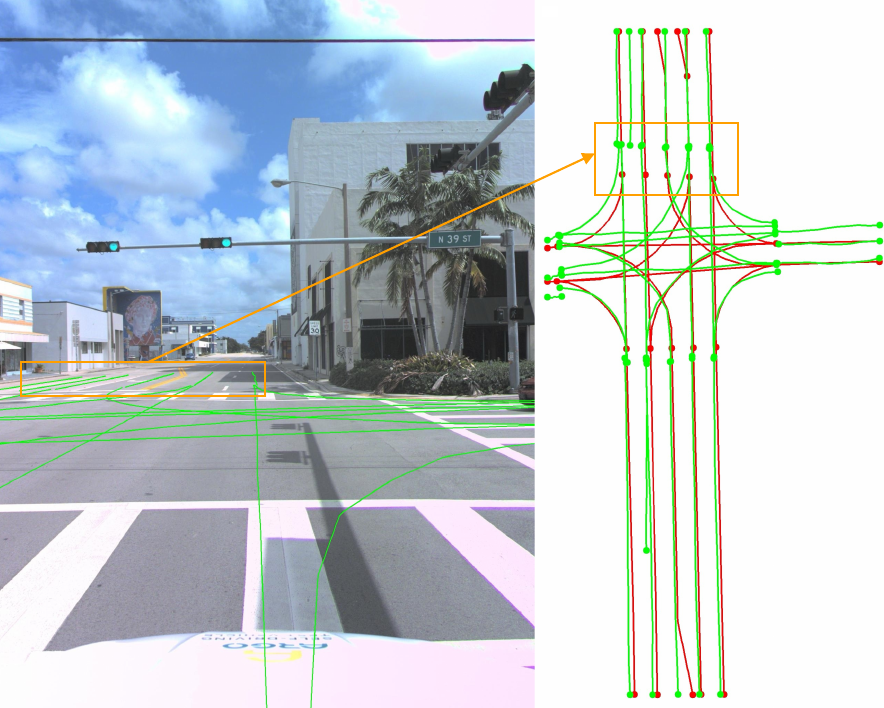}
    \caption{}
    \label{fig:centerlineb}
  \end{subfigure}
  \caption{\textbf{Visualization of 2D and 3D lane detection results} on OpenLane-V2 \textit{subset\_A}. 
  % (a) and (b) are two different scenarios. 
  (a) is an intersection scene and our method accurately detects the positions of all centerlines in this scene. (b) is a \emph{failure case} where our method predicts centerlines that align well with ground truths except for less precise starting/ending points.
  Ground truths are showed in \textcolor{Red}{red}, while predictions are showed in \textcolor{green}{green}. Best viewed in color.}
\end{figure}

\begin{figure}[t]
  \centering
  \includegraphics[width=0.92\linewidth]{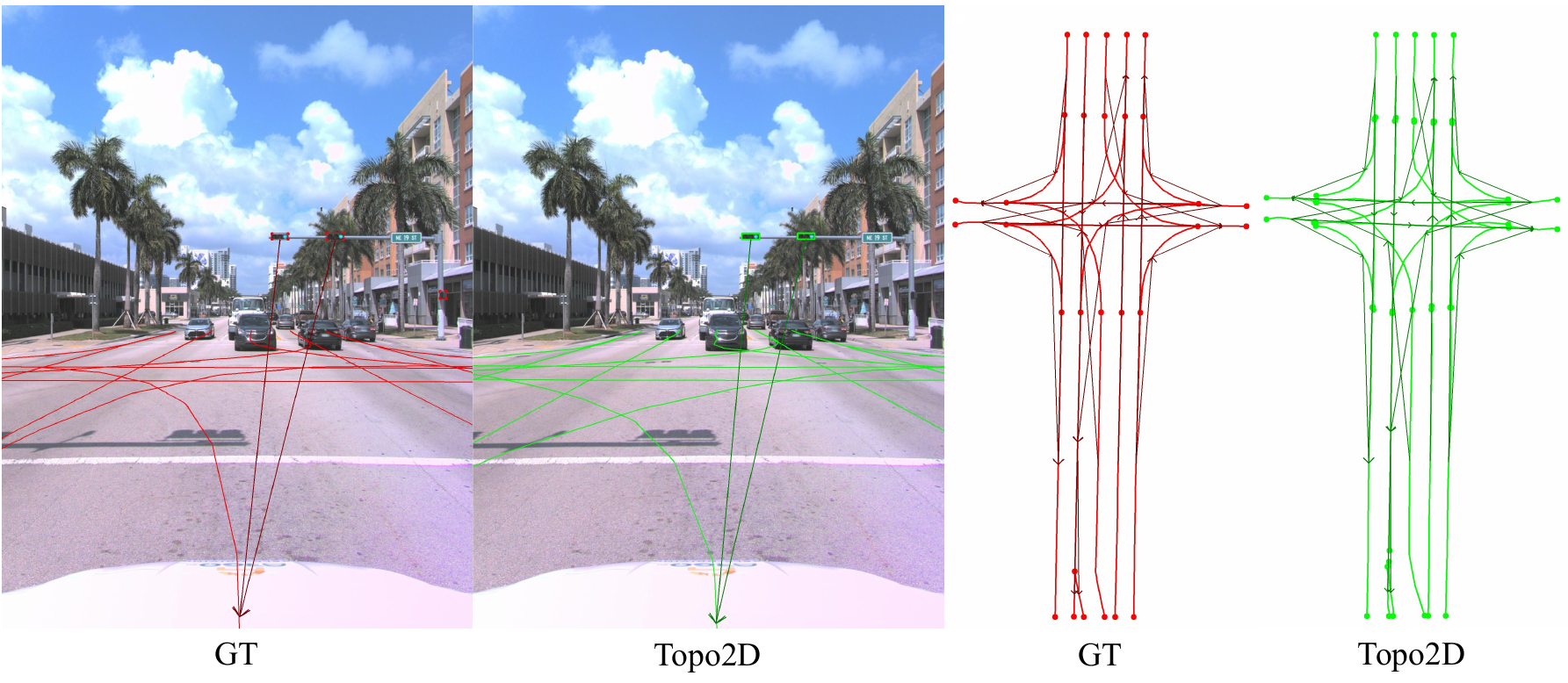}
  \caption{\textbf{Visualization of topology reasoning results} on OpenLane-V2 \textit{subset\_A}. Left: Lane-traffic element topology. Right: Lane-lane topology. Ground truths are showed in \textcolor{Red}{red}, while predictions are showed in \textcolor{green}{green}. Best viewed in color.}
  \label{fig:visopenlanev2topo}
\end{figure}

\begin{figure}[t]
  \centering
  \includegraphics[width=0.92\linewidth]{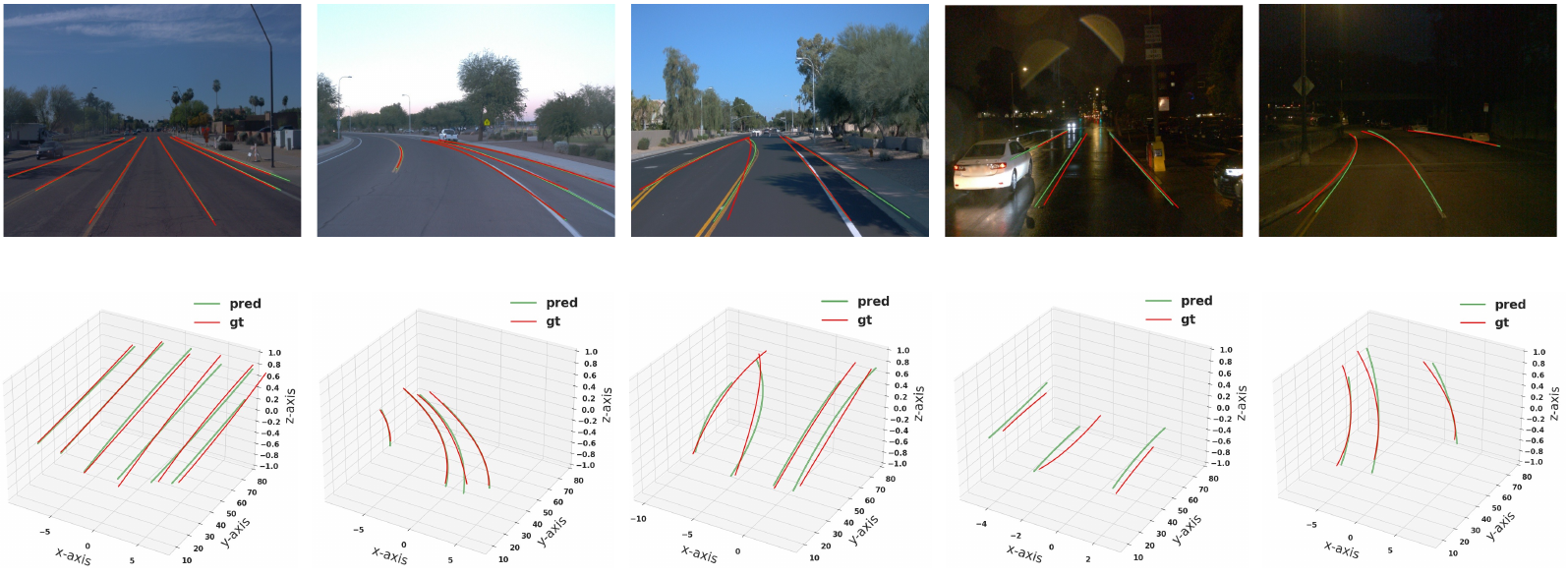}
  \caption{\textbf{Visualization under different scenarios} on OpenLane validation set.
  }
  \label{fig:visopenlane}
\end{figure}

\subsubsection{Initialization method for 3D queries.}
% 我们的方法使用两个6层的Transformer Decoder，并使用二维车道查询初始化三维车道查询，为了验证二维车道查询的作用，我们设计了只使用三维随机初始化车道查询，并将Transformer Decoder Layer设置为12层。可以发现，相比于只使用三维车道查询，我们的模型有较为明显的性能提升，验证了我们二维先验的作用。
% Our approach employs two 6-layer Transformer Decoders, initializing 3D lane queries with 2D lane queries. 
To evaluate the effectiveness of the 2D-based initialization, we conduct experiments employing various initialization methods for 3D lane queries. As demonstrated in \cref{tab:3dqueryinitial}, compared to random initialization in 3D space, our proposed 2D-based initialization method significantly enhances performance (28.8\% vs. 24.7\% on $\text{DET}_l$).
% 我们设计了在二维车道查询的基础上，添加部分随机初始化的三维查询的实验，性能相应的也会下降。我们发现随机初始化的三维车道查询只会匹配上相对简单的实例，而将较为困难的实例留给二维车道查询学习，这也验证了我们的motivation，二维查询更容易学习相对困难的实例。
Additionally, when mixing these two types of 3D lane queries while maintaining a constant number of queries, we observe a decline in performance. It is noted that 3D lane queries with random initialization tended to match only relatively simple instances, leaving the more complex instances to those queries initialized with 2D lane priors.
% This precisely validates our motivation.

\subsubsection{Number of instance queries.}
% 我们将每个视角的实例查询数量分别设置为10,20,30。由于数据集每个场景中的车道数量较多，查询数量设置为10时，模型的性能较低，设置成20和30时的性能相近。
In \cref{tab:querynum}, we configure the number of instance queries for each view to be 10, 20, and 30, respectively. Given the large number of lanes present in complex intersection scenes,  the model's performance declines when the query number is set to 10. Conversely, increasing the query number to 30 introduces a higher proportion of negative samples, which could negatively impact the training process, leading to slight performance degradation. Consequently, we set the query number to 20 as the default setting in our experiments.

\subsubsection{Design of 2D lane ground-truths.}
% 我们比较了二维车道检测器gt不同采样方式的性能，在使用2d sample方式时，我们截取每个视角中的可见部分，并进行等距采样。在使用3d sample方式时，我们将三维车道线进行等距采样，并投影到每个视角中。对于二维车道检测器，增加vis branch，预测当前视角该点是否可见。实验结果显示，采用2d sample方式，二维车道检测器的训练更加稳定，最后的检测结果也更好。
Considering that instances in the 3D scene may not entirely appear within a single 2D view, we compare the performance under different designs for 2D lane ground-truths.
When using the 2D sampling method, for each view, we project the visible parts of 3D lanes onto the image, and perform equidistant sampling on the image to generate 2D lane ground-truths.
When using the 3D sampling method, we equidistantly sample the entire 3D lanes in the 3D space and project all points onto the image. The 2D lane detector not only detects the point coordinates but also predicts their visibility status in current view. The differences of these two methods are more notable in farther distance.
% \textcolor{wzt}{don not understand...}
Experimental results in \cref{tab:2dsample} show that when using the 2D sample method, the training of the 2D lane detector is more stable, resulting in better detection performance.

% Through visualization and experimental analysis, we observed that the lane centerlines at intersections, lacking explicit ground markings as references and often having high curvature, are relatively difficult instances to learn and accurately predict their positions. These instances typically appear more completely in a single view. Our 2D lane queries concentrate on extracting features more comprehensively and globally for these lanes, leading to better predictions. Conversely, instances spanning multiple views, such as straight lines, are relatively easy to learn. 

\subsubsection{Instance query vs. hierarchical query.}
% 我们尝试将基于点的二维车道查询通过mlp融合成基于实例的三维车道查询，每个查询预测车道上所有点的三维坐标。实验结果表明，这种融合方式会导致三维车道查询在进行全局注意力机制时难以定位相应的图像特征，导致性能欠佳。
Before detecting 3D lanes, we also attempt to fuse hierarchical 2D lane queries into instance 3D lane queries using MLPs: 
\begin{equation}
  \mathbf{Q}_{i}^{3d}=\text{MLP}(\text{Concat}(\{\hat{\mathbf{Q}}_{ij}^{2d}\}_{j=1}^{N_P})).
  \label{eq:insquery}
\end{equation}
Each instance query predicts the coordinates of all points on this lane instance.
In \cref{tab:ptsquery}, experiment results indicate that this fusion approach makes it difficult for the 3D lane instance queries to locate the corresponding image features during cross attention module, resulting in sub-optimal performance. 

\subsubsection{Impact of adding 2D information.}
We evaluate the impact of adding 2D information 
% (e.g., 2D lane features) 
into the topology reasoning task.
% \textcolor{wzt}{(no positions?)} 
As illustrated in \cref{tab:2dquerytopo}, directly adding the projection matrix from 3D world coordinate to 2D image coordinate benefits the prediction of lane-traffic element topology. Introducing extra 2D query features results in further enhancements in both $\text{TOP}_{ll}$ and $\text{TOP}_{lt}$, verifying the value of adding 2D information for effective topology reasoning.
% since the detection of traffic elements is based on 2D images, directly adding the projection matrix of 3D world coordinate system to 2D image coordinate system benefits the prediction of lane-traffic element topology. 
% \textcolor{wzt}{how does projection matrix work?}
% With additional 2D lane queries features, 
% Furthermore, 2D lane queries aggregate image features that are more consistent with those of 2D traffic element queries. So adding 2D lane features brings obvious improvement to $\text{TOP}_{lt}$, verifying the effectiveness of our method.

\subsection{Visualization}
\cref{fig:centerlinea} and \cref{fig:visopenlanev2topo} show our lane detection and topology prediction results on OpenLane-V2\cite{Openlane-v2}. 
% On the left are the outputs of our 2D lane detector, while on the right are the outputs of our 3D lane detector (\textcolor{red}{red} lines represent GTs, \textcolor{green}{green} lines represent predictions). 
It can be observed that even in intersections without visual lane cues, our 2D lane detector still detect most of the centerlines, providing high-recall lane candidates for the 3D lane detector.
We also visualize the results on the OpenLane\cite{Persformer} dataset, as shown in \cref{fig:visopenlane}. 
% Topo2D achieve accurate detection results in various scenarios such as Curve, Merge, etc. 
% Especially in the Night scenes where image features are difficult to identify, the visualization results demonstrate that our 2D lane priors benefit the 3D detector in accurately locating lane image features.
More qualitative results are included in the supplementary materials.

\section{Conclusion}
This paper proposes a new framework Topo2D for lane detection and topology reasoning. By initializing 3D lane queries based on 2D lane priors, the 3D lane detector learns more comprehensive image features, achieving higher detection recall rate.
We further explicitly incorporate 2D lane features into the prediction of the topological structure.
Experimental results show that our Topo2D outperforms previous state-of-the-art methods on multi-view topology reasoning benchmark OpenLane-V2 and single-view 3D lane detection benchmark OpenLane.
% \subsubsection{Limitations.} 
% % During the initialization of 3D lane queries with 2D lane priors, the number of lane proposals provided from each view remains consistent. However, in certain perspective views where lane lines are absent, an excessive number of 2D false positive features might hinder the model's convergence speed.
% Inaccurate prediction of the starting and ending points of distant 2D lanes could cause a performance gap between $\text{DET}_l$ and $\text{DET}_{l,chamfer}$. We aim to solve this issue to have a more consistent detection performance in future research.

\bibliographystyle{splncs04}
\bibliography{main}

\begin{thebibliography}{10}
\providecommand{\url}[1]{\texttt{#1}}
\providecommand{\urlprefix}{URL }
\providecommand{\doi}[1]{https://doi.org/#1}

\bibitem{Curveformer}
Bai, Y., Chen, Z., Fu, Z., Peng, L., Liang, P., Cheng, E.: Curveformer: 3d lane detection by curve propagation with curve queries and attention. In: 2023 IEEE International Conference on Robotics and Automation (ICRA). pp. 7062--7068. IEEE (2023)

\bibitem{nuscenes}
Caesar, H., Bankiti, V., Lang, A.H., Vora, S., Liong, V.E., Xu, Q., Krishnan, A., Pan, Y., Baldan, G., Beijbom, O.: nuscenes: A multimodal dataset for autonomous driving. In: Proceedings of the IEEE/CVF conference on computer vision and pattern recognition. pp. 11621--11631 (2020)

\bibitem{stsu}
Can, Y.B., Liniger, A., Paudel, D.P., Van~Gool, L.: Structured bird's-eye-view traffic scene understanding from onboard images. In: Proceedings of the IEEE/CVF International Conference on Computer Vision. pp. 15661--15670 (2021)

\bibitem{detr}
Carion, N., Massa, F., Synnaeve, G., Usunier, N., Kirillov, A., Zagoruyko, S.: End-to-end object detection with transformers. In: European conference on computer vision. pp. 213--229. Springer (2020)

\bibitem{Persformer}
Chen, L., Sima, C., Li, Y., Zheng, Z., Xu, J., Geng, X., Li, H., He, C., Shi, J., Qiao, Y., et~al.: Persformer: 3d lane detection via perspective transformer and the openlane benchmark. In: European Conference on Computer Vision. pp. 550--567. Springer (2022)

\bibitem{gkt}
Chen, S., Cheng, T., Wang, X., Meng, W., Zhang, Q., Liu, W.: Efficient and robust 2d-to-bev representation learning via geometry-guided kernel transformer. arXiv preprint arXiv:2206.04584  (2022)

\bibitem{pivotnet}
Ding, W., Qiao, L., Qiu, X., Zhang, C.: Pivotnet: Vectorized pivot learning for end-to-end hd map construction. In: Proceedings of the IEEE/CVF International Conference on Computer Vision. pp. 3672--3682 (2023)

\bibitem{3d-lanenet}
Garnett, N., Cohen, R., Pe'er, T., Lahav, R., Levi, D.: 3d-lanenet: end-to-end 3d multiple lane detection. In: Proceedings of the IEEE/CVF International Conference on Computer Vision. pp. 2921--2930 (2019)

\bibitem{Gen-lanenet}
Guo, Y., Chen, G., Zhao, P., Zhang, W., Miao, J., Wang, J., Choe, T.E.: Gen-lanenet: A generalized and scalable approach for 3d lane detection. In: Computer Vision--ECCV 2020: 16th European Conference, Glasgow, UK, August 23--28, 2020, Proceedings, Part XXI 16. pp. 666--681. Springer (2020)

\bibitem{resnet}
He, K., Zhang, X., Ren, S., Sun, J.: Deep residual learning for image recognition. In: Proceedings of the IEEE conference on computer vision and pattern recognition. pp. 770--778 (2016)

\bibitem{Anchor3dlane}
Huang, S., Shen, Z., Huang, Z., Ding, Z.h., Dai, J., Han, J., Wang, N., Liu, S.: Anchor3dlane: Learning to regress 3d anchors for monocular 3d lane detection. In: Proceedings of the IEEE/CVF Conference on Computer Vision and Pattern Recognition. pp. 17451--17460 (2023)

\bibitem{Hdmapnet}
Li, Q., Wang, Y., Wang, Y., Zhao, H.: Hdmapnet: An online hd map construction and evaluation framework. In: 2022 International Conference on Robotics and Automation (ICRA). pp. 4628--4634. IEEE (2022)

\bibitem{toponet}
Li, T., Chen, L., Geng, X., Wang, H., Li, Y., Liu, Z., Jiang, S., Wang, Y., Xu, H., Xu, C., et~al.: Topology reasoning for driving scenes. arXiv preprint arXiv:2304.05277  (2023)

\bibitem{linecnn}
Li, X., Li, J., Hu, X., Yang, J.: Line-cnn: End-to-end traffic line detection with line proposal unit. IEEE Transactions on Intelligent Transportation Systems  \textbf{21}(1),  248--258 (2019)

\bibitem{Bevformer}
Li, Z., Wang, W., Li, H., Xie, E., Sima, C., Lu, T., Qiao, Y., Dai, J.: Bevformer: Learning bird’s-eye-view representation from multi-camera images via spatiotemporal transformers. In: European conference on computer vision. pp. 1--18. Springer (2022)

\bibitem{Maptr}
Liao, B., Chen, S., Wang, X., Cheng, T., Zhang, Q., Liu, W., Huang, C.: Maptr: Structured modeling and learning for online vectorized hd map construction. arXiv preprint arXiv:2208.14437  (2022)

\bibitem{Maptrv2}
Liao, B., Chen, S., Zhang, Y., Jiang, B., Zhang, Q., Liu, W., Huang, C., Wang, X.: Maptrv2: An end-to-end framework for online vectorized hd map construction. arXiv preprint arXiv:2308.05736  (2023)

\bibitem{fpn}
Lin, T.Y., Doll{\'a}r, P., Girshick, R., He, K., Hariharan, B., Belongie, S.: Feature pyramid networks for object detection. In: Proceedings of the IEEE conference on computer vision and pattern recognition. pp. 2117--2125 (2017)

\bibitem{focal}
Lin, T.Y., Goyal, P., Girshick, R., He, K., Doll{\'a}r, P.: Focal loss for dense object detection. In: Proceedings of the IEEE international conference on computer vision. pp. 2980--2988 (2017)

\bibitem{Vectormapnet}
Liu, Y., Yuan, T., Wang, Y., Wang, Y., Zhao, H.: Vectormapnet: End-to-end vectorized hd map learning. In: International Conference on Machine Learning. pp. 22352--22369. PMLR (2023)

\bibitem{petr}
Liu, Y., Wang, T., Zhang, X., Sun, J.: Petr: Position embedding transformation for multi-view 3d object detection. In: European Conference on Computer Vision. pp. 531--548. Springer (2022)

\bibitem{Petrv2}
Liu, Y., Yan, J., Jia, F., Li, S., Gao, A., Wang, T., Zhang, X.: Petrv2: A unified framework for 3d perception from multi-camera images. In: Proceedings of the IEEE/CVF International Conference on Computer Vision. pp. 3262--3272 (2023)

\bibitem{cos}
Loshchilov, I., Hutter, F.: Sgdr: Stochastic gradient descent with warm restarts. arXiv preprint arXiv:1608.03983  (2016)

\bibitem{adamw}
Loshchilov, I., Hutter, F.: Decoupled weight decay regularization. arXiv preprint arXiv:1711.05101  (2017)

\bibitem{Latr}
Luo, Y., Zheng, C., Yan, X., Kun, T., Zheng, C., Cui, S., Li, Z.: Latr: 3d lane detection from monocular images with transformer. In: Proceedings of the IEEE/CVF International Conference on Computer Vision. pp. 7941--7952 (2023)

\bibitem{dice}
Milletari, F., Navab, N., Ahmadi, S.A.: V-net: Fully convolutional neural networks for volumetric medical image segmentation. In: 2016 fourth international conference on 3D vision (3DV). pp. 565--571. Ieee (2016)

\bibitem{vpn}
Pan, B., Sun, J., Leung, H.Y.T., Andonian, A., Zhou, B.: Cross-view semantic segmentation for sensing surroundings. IEEE Robotics and Automation Letters  \textbf{5}(3),  4867--4873 (2020)

\bibitem{qin2020ultra}
Qin, Z., Wang, H., Li, X.: Ultra fast structure-aware deep lane detection. In: Computer Vision--ECCV 2020: 16th European Conference, Glasgow, UK, August 23--28, 2020, Proceedings, Part XXIV 16. pp. 276--291. Springer (2020)

\bibitem{shan2018lego}
Shan, T., Englot, B.: Lego-loam: Lightweight and ground-optimized lidar odometry and mapping on variable terrain. In: 2018 IEEE/RSJ International Conference on Intelligent Robots and Systems (IROS). pp. 4758--4765. IEEE (2018)

\bibitem{shan2020lio}
Shan, T., Englot, B., Meyers, D., Wang, W., Ratti, C., Rus, D.: Lio-sam: Tightly-coupled lidar inertial odometry via smoothing and mapping. In: 2020 IEEE/RSJ international conference on intelligent robots and systems (IROS). pp. 5135--5142. IEEE (2020)

\bibitem{gridmask}
Singh, K.K., Yu, H., Sarmasi, A., Pradeep, G., Lee, Y.J.: Hide-and-seek: A data augmentation technique for weakly-supervised localization and beyond. arXiv preprint arXiv:1811.02545  (2018)

\bibitem{Waymo}
Sun, P., Kretzschmar, H., Dotiwalla, X., Chouard, A., Patnaik, V., Tsui, P., Guo, J., Zhou, Y., Chai, Y., Caine, B., et~al.: Scalability in perception for autonomous driving: Waymo open dataset. In: Proceedings of the IEEE/CVF conference on computer vision and pattern recognition. pp. 2446--2454 (2020)

\bibitem{laneatt}
Tabelini, L., Berriel, R., Paixao, T.M., Badue, C., De~Souza, A.F., Oliveira-Santos, T.: Keep your eyes on the lane: Real-time attention-guided lane detection. In: Proceedings of the IEEE/CVF conference on computer vision and pattern recognition. pp. 294--302 (2021)

\bibitem{transformer}
Vaswani, A., Shazeer, N., Parmar, N., Uszkoreit, J., Jones, L., Gomez, A.N., Kaiser, {\L}., Polosukhin, I.: Attention is all you need. Advances in neural information processing systems  \textbf{30} (2017)

\bibitem{Openlane-v2}
Wang, H., Li, T., Li, Y., Chen, L., Sima, C., Liu, Z., Wang, B., Jia, P., Wang, Y., Jiang, S., et~al.: Openlane-v2: A topology reasoning benchmark for unified 3d hd mapping. Advances in Neural Information Processing Systems  \textbf{36} (2024)

\bibitem{bevlanedet}
Wang, R., Qin, J., Li, K., Li, Y., Cao, D., Xu, J.: Bev-lanedet: An efficient 3d lane detection based on virtual camera via key-points. In: Proceedings of the IEEE/CVF Conference on Computer Vision and Pattern Recognition. pp. 1002--1011 (2023)

\bibitem{argoverse}
Wilson, B., Qi, W., Agarwal, T., Lambert, J., Singh, J., Khandelwal, S., Pan, B., Kumar, R., Hartnett, A., Pontes, J.K., et~al.: Argoverse 2: Next generation datasets for self-driving perception and forecasting. arXiv preprint arXiv:2301.00493  (2023)

\bibitem{TopoMLP}
Wu, D., Chang, J., Jia, F., Liu, Y., Wang, T., Shen, J.: Topomlp: An simple yet strong pipeline for driving topology reasoning. arXiv preprint arXiv:2310.06753  (2023)

\bibitem{clrnet}
Zheng, T., Huang, Y., Liu, Y., Tang, W., Yang, Z., Cai, D., He, X.: Clrnet: Cross layer refinement network for lane detection. In: Proceedings of the IEEE/CVF conference on computer vision and pattern recognition. pp. 898--907 (2022)

\bibitem{cvt}
Zhou, B., Kr{\"a}henb{\"u}hl, P.: Cross-view transformers for real-time map-view semantic segmentation. In: Proceedings of the IEEE/CVF conference on computer vision and pattern recognition. pp. 13760--13769 (2022)

\bibitem{deformdetr}
Zhu, X., Su, W., Lu, L., Li, B., Wang, X., Dai, J.: Deformable detr: Deformable transformers for end-to-end object detection. arXiv preprint arXiv:2010.04159  (2020)

\end{thebibliography}

\clearpage
\appendix
\noindent{\Large \textbf{Appendix}}

\section{More Implementation Details}
For the 2D lane detector, in order to adapt to the arbitrary shape of 2D lanes, we add an edge direction loss\cite{Maptr} to supervise the geometrical shape with a loss weight set to 0.005. 
In addition, we add an auxiliary segmentation head to predict 2D lane instance masks on OpenLane\cite{Persformer} dateset. The segmentation loss is defined as:
\begin{equation}
  \mathcal{L}_{\text{seg}} = \mathcal{L}_{\text{obj}} + \mathcal{L}_{\text{dice}} + \mathcal{L}_{\text{pixel}},
  \label{eq:segloss}
\end{equation}
where $\mathcal{L}_{\text{obj}}$, $\mathcal{L}_{\text{dice}}$ and $\mathcal{L}_{\text{pixel}}$ are binary cross entropy loss for the IoU-aware objectness, dice loss\cite{dice} and pixel-wise binary cross entropy loss for segmentation mask, respectively. The loss weights for $\mathcal{L}_{\text{obj}}$, $\mathcal{L}_{\text{dice}}$ and $\mathcal{L}_{\text{pixel}}$ are set to 1.0, 2.0 and 5.0.
For the predictions of the lane detector and traffic element detector, we use bipartite matching to assign ground truths. This matching results are directly used for topology reasoning loss.

For the OpenLane-V2 dataset\cite{Openlane-v2}, the front view images are first cropped and padded to match the size of other views. Then all images are resized to $775\times 1024$ with a scaling factor of 0.5. 
The batch size is 8 with an initial learning rate of 3e-4.
For the OpenLane\cite{Persformer} dataset, all input images are resized to $800\times 1024$. The structure of the lane detection model is similar to the centerline detection model used on OpenLane-V2\cite{Openlane-v2}. The batch size is 32 with an initial learning rate of 2e-4. 

Our 2D-based initialization method utilizes a 6-layer 2D lane decoder and a 6-layer 3D lane decoder. In the ablation experiments exploring various initialization methods for 3D lane queries, to ensure fairness in comparison, we set the 3D lane decoder to 12 layers when employing the random initialization method.

\section{Quantitative Results}

\subsection{Comparison on OpenLane-V2 Dataset}
We provide the performance of topology reasoning task using metrics before updating. 
As depicted in \cref{tab:openlanev2sotatopo}, our method achieves 37.8\% OLS, indicating a 2.2\% improvement compared to TopoNet\cite{toponet}.

\subsection{Comparison on OpenLane Dataset}
We provide the performance comparison under different scenarios in \cref{tab:openlanef1}.
Topo2D outperforms LATR\cite{Latr} across most scenarios. 
Specifically, we observe that our model performs more accurate lane detection under scenarios such as Extreme Weather and Night, where identifying 2D lane features becomes particularly challenging.
This is due to our 2D lane queries extracting more comprehensive image information, allowing the 3D queries based on 2D lane prior to better locate lane features in the 3D lane decoder, achieving more accurate predictions of lane point positions.

\section{Qualitative Results}

\begin{table}[tb]
  \caption{\textbf{Comparison on topology reasoning task} on OpenLane-V2 \textit{subset\_A}. 
  The reported results of state-of-art methods are from TopoNet\cite{toponet}. 
  % The best is in \textbf{bold}.
  }
  \label{tab:openlanev2sotatopo}
  \centering
  \begin{tabular}{@{}l|cc|c|cccc@{}}
    \toprule 
    Method & Backbone & Epoch & $\text{OLS}\uparrow$ & $\text{DET}_l\uparrow$ & $\text{DET}_t\uparrow$ & $\text{TOP}_{ll}\uparrow$ & $\text{TOP}_{lt}\uparrow$ \\
    \midrule
    STSU\cite{stsu} & ResNet-50 & 24e & 25.4 & 12.7 & 43.0 & 0.5 & 15.1 \\
    VectorMapNet\cite{Vectormapnet}$\;$ & ResNet-50 & 24e & 20.8 & 11.1 & 41.7 & 0.4 & 5.9 \\
    MapTR\cite{Maptr} & ResNet-50 & 24e & 26.0 & 17.7 & 43.5 & 1.1 & 10.4 \\
    % paper
    % TopoNet\cite{toponet} & ResNet-50 & 24e & 35.6 & 28.5 & 48.1 & 4.1 & 20.8 \\
    % github
    TopoNet\cite{toponet} & ResNet-50 & 24e & 35.6 & 28.6 & 48.6 & 4.1 & 20.3 \\
    % paper
    % TopoMLP\cite{TopoMLP} & ResNet-50 & 24e & 38.2 & 28.3 & 50.0 & 7.2 & 22.8 \\
    % github
    % TopoMLP\cite{TopoMLP} & ResNet-50 & 24e & 38.2 & 28.5 & 49.5 & 7.2 & 23.4 \\
    Topo2D (Ours) & ResNet-50 & 24e & \textbf{37.8} & \textbf{29.1} & \textbf{50.6} & \textbf{6.6} & \textbf{21.1} \\
  \bottomrule
  \end{tabular}
\end{table}

\begin{table}[tb]
  \caption{\textbf{Comparison with state-of-art methods} on OpenLane validation set.}
  \label{tab:openlanef1}
  \centering
  \begin{tabular}{@{}l|>{\centering}p{1.0cm}|>{\centering}p{1.0cm}|>{\centering}p{1.0cm}|>{\centering}p{1.25cm}|>{\centering}p{1.0cm}|>{\centering}p{1.75cm}|c@{}}
    \toprule 
    & & Up \& & & Extreme & & & Merge \\
    \multirow{-2}*{Method} & \multirow{-2}*{All} & Down & \multirow{-2}*{Curve} & Weather & \multirow{-2}*{Night} & \multirow{-2}*{Intersection} & \& Split \\
    \midrule
    3D-LaneNet\cite{3d-lanenet} & 44.1 & 40.8 & 46.5 & 47.5 & 41.5 & 32.1 & 41.7 \\
    Gen-LaneNet\cite{Gen-lanenet} & 32.3 & 25.4 & 33.5 & 28.1 & 18.7 & 21.4 & 31.0 \\
    PersFormer\cite{Persformer} & 50.5 & 42.4 & 55.6 & 48.6 & 46.6 & 40.0 & 50.7 \\
    CurveFormer\cite{Curveformer} & 50.5 & 45.2 & 56.6 & 49.7 & 49.1 & 42.9 & 45.4 \\
    Anchor3DLane\cite{Anchor3dlane}$\;$ & 53.7 & 46.7 & 57.2 & 52.5 & 47.8 & 45.4 & 51.2 \\
    BEVLaneDet\cite{bevlanedet} & 58.4 & 48.7 & 63.1 & 53.4 & 53.4 & 50.3 & 53.7 \\
    LATR\cite{Latr} & 61.9 & 55.2 & \textbf{68.2} & 57.1 & 55.4 & 52.3 & 61.5 \\
    Topo2D (Ours) & \textbf{62.6} & \textbf{55.5} & 67.7 & \textbf{59.1} & \textbf{57.4} & \textbf{52.4} & \textbf{62.5} \\
  \bottomrule
  \end{tabular}
\end{table}

\begin{figure}[tb]
  \centering
  \includegraphics[width=0.8\linewidth]{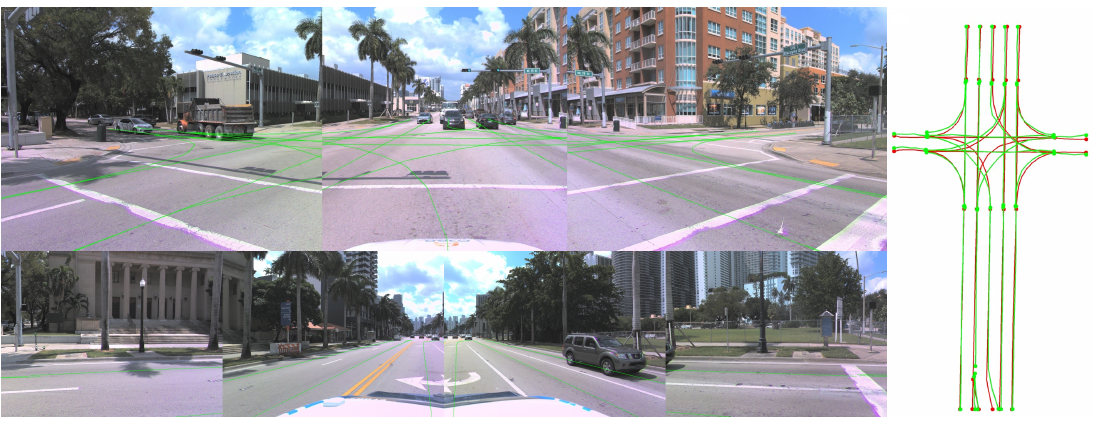}
  \caption{\textbf{Visualization of 2D and 3D lane detection results} on OpenLane-V2 \textit{subset\_A}. Ground truths are showed in \textcolor{Red}{red}, while predictions are showed in \textcolor{green}{green}. Best viewed in color.}
  \label{fig:vis2dlane}
\end{figure}

\begin{figure}[tb]
  \centering
  \includegraphics[width=0.8\linewidth]{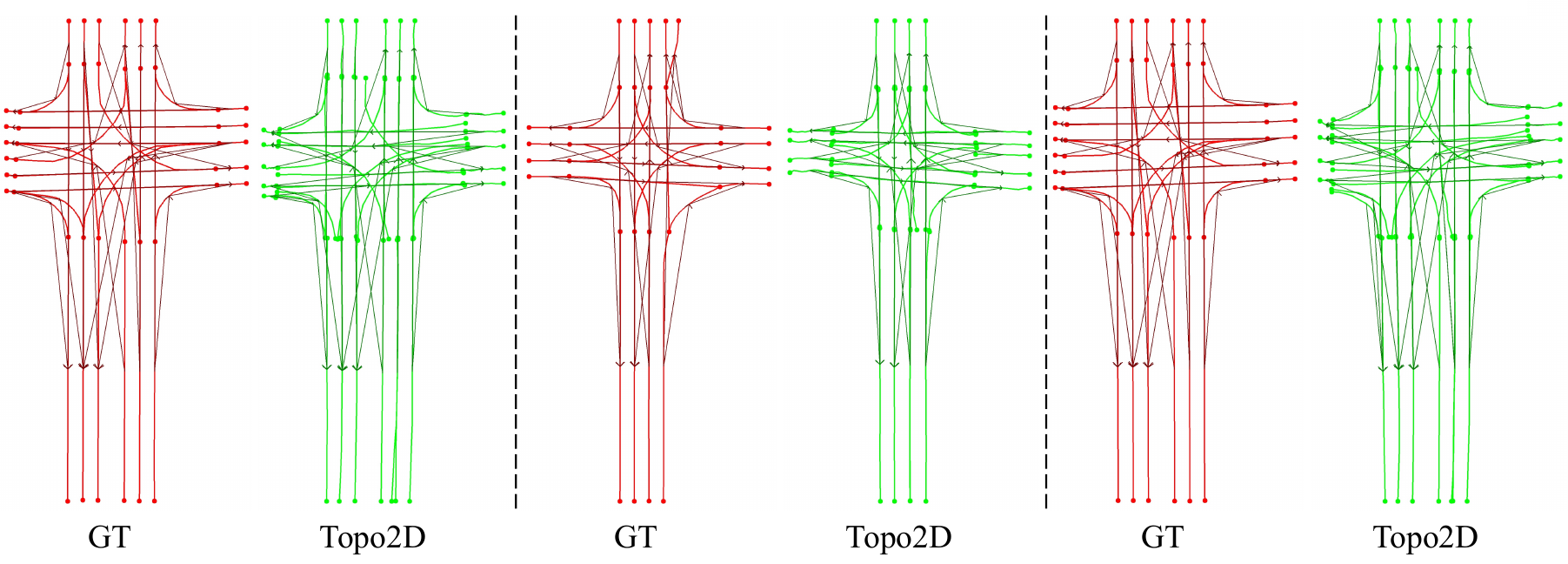}
  \caption{\textbf{Visualization of lane-lane topology reasoning results} on OpenLane-V2 \textit{subset\_A}. Ground truths are showed in \textcolor{Red}{red}, while predictions are showed in \textcolor{green}{green}. Best viewed in color.}
  \label{fig:vistopo}
\end{figure}

\begin{figure}[tb]
  \centering
  \includegraphics[width=0.8\linewidth]
  {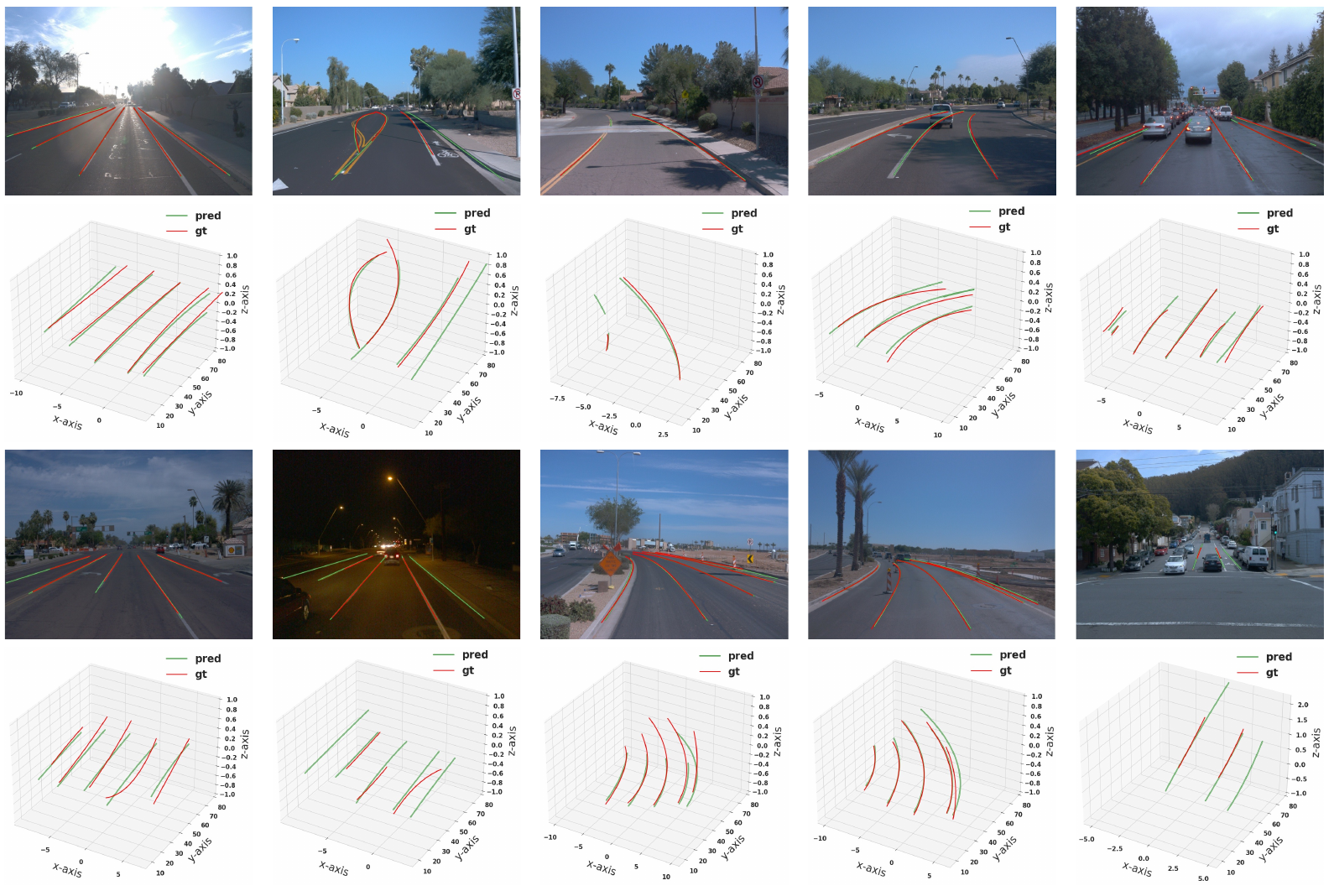}
  \caption{\textbf{Visualization under different scenarios} on OpenLane validation set.
  }
  \label{fig:visopenlanemore}
\end{figure}

For the OpenLane-V2\cite{Openlane-v2} dateset, we visualize the 2D lane detection results for all camera views in \cref{fig:vis2dlane}. 
Additionally, to validate the robust performance of our method in topology reasoning task, we visualize lane-lane topology prediction results across more diverse scenarios in \cref{fig:vistopo}. It can be observed that even in highly complex scenarios, Topo2D consistently provides accurate predictions of road topology structures.
For the OpenLane\cite{Persformer} dataset, more visualization results are presented in \cref{fig:visopenlanemore}. 

\end{document}